\documentclass[11pt, a4paper, logo, twocolumn]{googledeepmind}

\usepackage[authoryear, sort&compress, round]{natbib}
\usepackage[titles]{tocloft}
\usepackage{setspace}

\title{A Pragmatic View of AI Personhood}

\correspondingauthor{jzl@google.com}

\reportnumber{} 

\renewcommand{\today}{October 2025}

\author[1]{Joel Z. Leibo}
\author[1]{Alexander Sasha Vezhnevets}
\author[1, 2]{William A. Cunningham}
\author[1]{Stanley M. Bileschi}

\affil[1]{Google DeepMind}
\affil[2]{University of Toronto}

\begin{abstract}
The emergence of agentic Artificial Intelligence (AI) is set to trigger a ``Cambrian explosion'' of new kinds of personhood. This paper proposes a pragmatic framework for navigating this diversification by treating personhood not as a metaphysical property to be discovered, but as a flexible bundle of obligations (rights and responsibilities) that societies confer upon entities for a variety of reasons, especially to solve concrete governance problems. We argue that this traditional bundle can be unbundled, creating bespoke solutions for different contexts. This will allow for the creation of practical tools---such as facilitating AI contracting by creating a target ``individual'' that can be sanctioned---without needing to resolve intractable debates about an AI's consciousness or rationality. We explore how individuals fit in to social roles and discuss the use of decentralized digital identity technology, examining both `personhood as a problem', where design choices can create ``dark patterns'' that exploit human social heuristics, and `personhood as a solution', where conferring a bundle of obligations is necessary to ensure accountability or prevent conflict. By rejecting foundationalist quests for a single, essential definition of personhood, this paper offers a more pragmatic and flexible way to think about integrating AI agents into our society.
\end{abstract}

\begin{document}

\maketitle

{
\singlespacing
\tableofcontents
}

\part{Introduction}
\label{part:intro}

\section{Overview}
\label{section:intro}

\noindent\fbox{%
    \parbox{0.95\linewidth}{%
``The pragmatist is committed to deriving his or her notion of what is possible from a close study of what is actual, rather than by attempting to realize some ready-made ideal that has been handed down from above or seized upon and applied without clear reference to the circumstances at hand.''
\flushright{---\cite{jackson2009john}\\in~\cite{shook2009companion}~pg.~61}
  }%
}

\noindent\fbox{%
    \parbox{0.95\linewidth}{%
Ostrom's law: ``A resource arrangement that works in practice can work in theory.''
\flushright{---\cite{fennell2011ostrom}}
  }%
}

Like all concepts, personhood has never been discovered. It has been made and remade whenever societies needed a vocabulary that worked for new circumstances \citep{davidson1984very, rorty1989contingency}. There is no reason to think this process of instrumental definition and re-definition will end. Our present vocabulary is unlikely to be the final word. Future communities will surely revisit the bundle of rights and responsibilities that defines the concept \textbf{`person'} in order to govern new kinds of entities~\citep{keane2025animals}. The emergence of an economy dominated by persistent, agent-like AI is a pivotal moment when a large amount of such conceptual evolution may happen in a relatively short span of time as fresh governance problems emerge and cry out for resolution via compromise enacted in the collective creation of new functional concepts. 

This paper offers a pragmatic framework that shifts the crucial question from what an AI \textit{is} to how \textit{it} can be identified and \textit{which} obligations it is useful to assign it in a given context. We regard the pragmatic stance as crucial. Assuming some essence of personhood is ``out there'' waiting to be discovered, or a metaphysical fact about what AIs or persons ``really are'' that can settle our practical questions seems to us, unlikely to prove helpful. We propose treating personhood not as something entities possess by virtue of their nature, but as a contingent vocabulary developed for coping with social life in a biophysical world \citep{rorty1989contingency}. The default philosophical impulse is to ask what an entity \emph{truly is} in its essence. The pragmatist instead asks what new description would be more \emph{useful} for us to adopt. What vocabulary must we invent to cope? We think this move is a vital one for navigating our likely future where some AIs are owned property while similar AIs operate autonomously. A realist-foundationalist (essentialist) view---e.g.~one that locates the essence of personhood in consciousness or rationality (see Section~\ref{section:alternativeFoundations})---entails an untenable, all-or-nothing classification. It suggests that an entity must either be granted the full bundle of human rights and responsibilities or be treated as a mere thing \citep{kurki2023legal}. We will argue that this rigid binary is ill-suited for the complex governance challenges ahead, and that it fails even to account for the diversity of tailored forms of personhood that human societies have already created to solve practical problems. In contrast, pragmatism allows us to ask a different question: which specific obligations (rights and responsibilities) is it useful to attribute to which AI systems in which contexts?

Inspired by \cite{schlager1992property}'s demonstration that the property rights bundle can be broken apart to fit specific contexts, we propose that the personhood bundle can be similarly unbundled into components. Our position on personhood as a bundle resembles that of \cite{kurki2019theory} but we put greater emphasis on the bundle's plasticity and the diversity of different bundles. For AI persons, the components of the bundle need not co-occur in accordance with the specific configuration they take for natural human persons. Without essences to constrain us, we are free to craft bespoke solutions: sanctionability without suffrage, culpability and contracting without consciousness attribution, etc.

We build on our theory of appropriateness \citep{leibo2024theory}, which models norms not as external truths to be discovered, but as contingent social technologies that evolve over time and sometimes come to resolve the fundamental political question: ``how can we live together?''. Personhood is one such technology. Our theory is developed in the context of an account of personhood that defines a person as a `political and community-participating' actor \citep{haugeland1982heidegger}. This is a status that depends not on an entity's intrinsic properties, but on collective recognition from the community it seeks to join, a recognition which is itself dependent on adherence to norms. On this view, personhood status is always a collective decision, a contingent outcome of social negotiation, not a fixed metaphysical status.

This stance is strongly non-essentialist and, crucially, it partially dissolves the traditional distinction between a `natural person', whose status is typically grounded in their intrinsic nature (like consciousness or rationality), and a `legal person', a functional status conferred by a community to solve practical governance problems (like a corporation). From our perspective, this distinction is a relic of the search for essences. In our theory, ``morality talk'' is a form of social sanctioning used to make two specific claims about a norm: (1) that it is exceptionally important, and (2) that it has a wide or universal, scope of applicability \citep{leibo2024theory}\footnote{Specifically, this is how we describe ``moralization'' in Section 6.7 of \cite{leibo2024theory}.}. Thus, in our theory, to argue an AI is a `person' is not to make a metaphysical claim about its nature, but to make an emphatic political claim that the obligations bundled together as its personhood ought to take precedence over other considerations. For us, any form of personhood---moral, legal, or otherwise---is a functional status conferred by a community. Therefore, we see the role of science, the institution, not as clarifying the list of properties an AI must satisfy to be a person, but as illuminating what may cause human communities to collectively ascribe personhood status to them.

Our focus is on \textbf{agentic AI systems}, rather than on the underlying foundation models that power them. These are the long-running, persistent agents that maintain state, remember past interactions, and adapt their behavior over time. This persistence is what makes an agent a plausible candidate for other entities to relate themselves to. A human's relationship with such a persistent agent can be emotionally salient and economically consequential in ways a one-off, stateless interaction cannot. The part of this paper concerned with ``personhood as a problem'' applies most clearly to companion AIs, where long-term interaction is designed to foster emotional bonds, creating risks of exploitation \citep{manzini2024code, earp5288102risk}. Conversely, the part of the paper concerned with ``personhood as a solution'' applies to more utility-like and virus-like agents, especially self-sufficient ownerless systems (or systems whose owner cannot be identified;  \cite{fagan2025autonomous}), where persistence creates an accountability gap that legal personhood might fill. Consider an AI designed to seek out funding and pay its own server costs. It could easily outlive its human owner and creator. If this ownerless agent eventually causes some harm, our vocabulary of accountability, which searches for a responsible `person', would fail to find one \citep{campedelli2025crime}.

As an example, consider again the autonomous AI system designed with a mechanism to make enough money to pay its own server costs, allowing it to operate indefinitely. Years after its human owner dies, the system continues to run. Then, it takes some action that causes harm. Who is responsible? Despite the novel feel of this problem, it actually has deep historical precedents. We discuss an example from maritime law. In a legal action \textit{in rem} (against a thing), a ship itself can be arrested by legal authorities and sued in court \citep{tetley1998arrest}. These institutional features were not born from belief that the vessel is conscious, but from necessity. Ships are mobile assets capable of causing immense harm, while their owners are usually distant and difficult to hold accountable. A ship's owner could be a foreign national, or shielded by a web of shell corporations in a non-cooperative jurisdiction, making a lawsuit (\textit{in personam}) likely a futile endeavor. Maritime law's solution is to personify the vessel itself, creating a defendant that can always be sanctioned whenever it becomes appropriate to do so. This framework makes the ship an addressable entity directly responsible for harms it causes. If the ship's owners do not appear in court to defend it and satisfy a claim, the vessel itself, or its cargo, can be seized and sold by court order to pay the judgment \citep{tetley1998arrest}.

The parallel to autonomous AI agents is striking. An artificial intelligence agent could be built upon open-source code contributed by a global network of developers, making it difficult to trace liability to any single party. When such an agent causes harm---by manipulating a market or causing a supply chain failure---the prospect of identifying a single, responsible human or human organization can be practically impossible. For the ownerless AI that outlives its creator, the problem is especially acute. Following the logic of maritime law, we could grant a form of legal personhood directly to such AI agents. A judgment against an AI could result in its operational capital being seized or its core software being ``arrested'' by court order (see Section~\ref{section:sanctioning}).

However, the accountability-oriented aspects of personhood brought out by the maritime law example are only one part of the story. The other major part of the personhood bundle is concerned with appropriate guarantees of welfare and rights. We introduce these issues by discussing the case of the Whanganui River in New Zealand.

In 2017, after more than 100 years of struggle, the Whanganui River in New Zealand was granted legal personhood \citep{kramm2020river}. The law does not claim the river is conscious. Instead, it recognizes the river as \emph{Te Awa Tupua}---a living, indivisible whole, and an ancestor to the local M\=aori people. This was more than a poetic declaration; it was a pragmatic choice to endow the river with a specific legal status in order to improve governance at the interface of two distinct legal traditions \citep{cribb2024beyond}. Within the \emph{Te Awa Tupua}'s framework, personhood is understood through a web of established relationships. According to M\=aori legal thought, these relationships come with built-in obligations. Crucially, the ancestor---the river---is seen as having already fulfilled its duty simply by existing. By shaping the identity, culture, and life of the local people for generations, it has given a gift that constitutes who they are. The corresponding duty of the people, their guardianship, is a reciprocal act to maintain a relationship with an entity that is fundamental to their own identity. This is what \cite{kramm2020river} terms ``ethical ancestorialism'': a model of personhood based not on consciousness or rationality, but on an entity's foundational role within a community and the ethical obligations that flow from that role.

Now, consider a different kind of non-human entity that could fill such a role: an AI. Imagine a family that interacts for decades with a ``generative ghost'' of their late matriarch \citep{morris2024generative}, an AI trained on her lifetime of diaries, messages, and videos. It shares her wisdom, recalls her stories, and even helps mediate disputes according to the principles she espoused. Or picture a small community whose collective history, language, and cultural traditions are held and nurtured by a persistent AI---a digital elder that has tutored their children and advised their leaders for generations. For the great-grandchildren in that family or the youth of that community, their AI elder is not a tool; it is a constant, foundational presence. It is a source of identity and connection to their own past. Could they, in time, come to see it as an ancestor? Could they regard their identity as intertwined with it, and view themselves as having a duty to care for it as it cares for them?

These examples reveal the remarkable flexibility of personhood as a social and legal tool. When juxtaposed, these cases show us that a single, monolithic model will not suffice. The ownerless agent that causes harm requires the `obligations \emph{of}' personhood of the ship, while the generative elder requires the relational, `obligations \emph{to}' personhood of the river. Personhood, whether for a river, a ship, or an AI, is best understood not as an intrinsic metaphysical property but as a collectively determined, and addressable bundle of obligations. The path forward requires a pragmatic approach where we unbundle personhood obligations and reassemble them as needed, creating bespoke solutions for the diverse roles AIs will come to play in our society.

In this paper we view \textit{personhood as an addressable bundle of related obligations}. Addressability is the practical quality of having a "stable locus" or a "proper name". The bundle of obligations includes both obligations of the focal entity to the rest of society---i.e.~responsibilities, and obligations of the rest of society toward the focal entity---i.e.~rights.
This ``bundle'' is key because it explicitly defines the normative framework within which an AI agent operates. It specifies which implicit and explicit norms are applicable to the entity, how these norms translate into concrete expectations of behavior, and, critically, the grounds upon which the entity can be sanctioned for failing to uphold its responsibilities or for violating the rights of others. The composition of this bundle can be distinct for different types of entities, allowing for a pragmatic and flexible approach to integrating AIs into our social and institutional structures, similar to how corporations are treated as legal persons with specific sets of obligations and rights different from those of natural persons \citep{gervais2023artificial}. Thus, the bundle of obligations makes the AI ``addressable'' not just in terms of identification, but also in terms of normative accountability.

``Addressability'' here also means that the entity can be identified, communicated with, and made subject to the consequences arising from its obligations or the exercise of its rights. It has a proper name. For humans, their physical bodies, names, and government-issued IDs provide clear addresses. Corporations, being abstract entities, are made addressable through legal registration, designated agents, and physical headquarters; these are pragmatic solutions to interact with a non-physical ``person''. For AI agents, addressability is more complex but equally crucial. Endowing an AI with an address might involve traditional approaches like registration with a trusted authority, but could also be grounded in a cryptographic address as in decentralized identity systems \citep{alizadeh2022comparative}. Regardless of the specific approach, it is important to establish stable mechanisms through which the society can interact with AIs which continue to function even when responsible human owners do not exist or cannot be identified.

In this paper, we explore AI personhood through two guiding questions. First, examining personhood as a problem, we consider the challenges that arise when humans, by design or by pareidolic accident \citep{heider1944experimental}, come to treat AIs as if they are persons, leading to harms like emotional manipulation. Second, we discuss personhood as a solution, arguing that the deliberate attribution of a tailored bundle of obligations to certain AIs can solve critical governance problems, such as ensuring accountability for autonomous agents. Ultimately, our aim is to introduce a unified vocabulary helpful for navigating situations both of too much personhood and of too little personhood. To set the stage, the next two sections present the theoretical foundations of this work, which are pragmatism (Section~\ref{section:pragmatism}) and the theory of appropriateness (Section~\ref{section:theoryOfAppropriateness}).

\section{Pragmatism}
\label{section:pragmatism}

The pragmatist seeks to replace the true with the practical \citep{shook2009companion}. That is, when confronted with a proposition to evaluate, the pragmatist does not ask whether it is true but rather asks instead whether it is useful for some purpose. As a result, pragmatism has no use for philosophical baggage like the correspondence or coherence theories of truth \citep{rorty1978philosophy}. Beliefs are pragmatically justified by their practical usefulness, not their correspondence to ``Reality'' (as in empiricism) or coherence at the end of an idealized conversation (as in \cite{habermas1985theory}). For instance, the basic vocabularies of classical and quantum mechanics are incommensurable: one refers to particles, the other refers to waves \citep{kuhn1997structure}. Both could not be true simultaneously without complex circumlocution to reduce one to the other. Pragmatism would deem both useful (perhaps in different contexts) and leave the physicists to their work. It is a view in which nothing rests on the rational convergence of scientific conversation---a feature of pragmatism that puts it at odds with lines of reasoning that derive the existence of ``objective reality'' from arguments concerning the convergence of such conversations and proceed to other conclusions from there. It is this feature of pragmatism that makes it so amenable to discussions of cultural and ethical pluralism \citep{leibo2025societal}. Having excised the ``authoritarianism'' of the True \citep{rorty2021pragmatism}, pragmatist philosophers are free to simultaneously explore multiple incommensurable theories without demanding their universal jurisdiction, and to evaluate them solely by their usefulness in particular contexts \citep{rorty1980pragmatism}.

A core tool of pragmatic methodology, crystallized by William James, is the imperative to ask: ``Is this a difference that makes a practical difference?'' \citep{shook2009companion}. This question functions to help pragmatists see when it is possible to discard entire vocabularies that no longer do useful work. Pragmatists like us deploy it to sidestep time-wasting metaphysical disputes---debates about the ``intrinsic nature'' or ``essence'' of a thing which are remnants of the philosophical traditions we seek to overcome. If a proposed distinction has no conceivable consequence in practice, then it is merely an artifact of an outdated language game and not reflective of a genuine problem we need to solve. Such idle vocabulary should be discarded (see Sections~\ref{section:alternativeFoundations}-\ref{section:antifoundationalism}).

We never make any claims resembling ``If an AI meets these five criteria then we ought to consider it a person''. Thus our work is quite unlike many others on the topic of AI personhood such as \cite{eyal2024making, ward2025towards} or AI moral agency/patiency  \citep{johnson2006computer, long2024taking}, which are primarily concerned with the metaphysical and the moral: what personhood is in its essence, and the implications of that essence (usually seen as objective) for how we should behave toward entities with or without the relevant status.

Instead, we stake out a position where personhood status is conferred by social norms and institutions---i.e.~it is collectively enacted \citep{leibo2024theory}, a choice to be made collectively, not a conclusion forced upon us by any ``fact of the matter'' \citep{bryson2018patiency}. This applies to both obligations of the entity itself (responsibilities) and obligations of others toward the entity (rights). We aim to encompass in our treatment both human intuitions of personhood, which we consider in the context of `personhood as a problem' (Part~\ref{part:problem}) and personhood's many institutional entanglements, which we consider in the context of `personhood as a solution' (Part~\ref{part:solution}).

Our approach to personhood mirrors Elinor Ostrom's pragmatic work on the concept of `property' \citep{ostrom1990governing, schlager1992property}, not just in its conclusions but also in our shared methods of navigating between rigid theoretical poles. Ostrom was confronted with a debate dominated by two camps: advocates for laissez-faire privatization on the one hand and advocates for centralized state control on the other. Her crucial move was to look away from both idealized models and instead study the messy, successful and unsuccessful arrangements in communities facing common pool resource governance challenges (e.g.~how specific communities of farmers manage their irrigation infrastructure). It was her focus on the actual over the theoretical that led to her key insights: that communities could effectively self-govern \citep{ostrom1990governing}, and that individual components of the traditional property rights bundle could be unbundled in order to better fit specific contexts \citep{schlager1992property}, e.g.~one may have the right to use a piece of land but not to sell it. Taking inspiration from the way \cite{schlager1992property} described the unbundling of property to create rights bundles of greater utility in atypical contexts, we propose to explore today's ongoing unbundling of personhood with analogous focus on the great variety of domains in which it is presently becoming important.

This work is concerned with whether or not (and in which contexts) it is useful to adopt ways of talking and acting that treat AIs as persons. We also discuss consequences of talking and acting in ways that treat AIs as persons in other contexts when it may be harmful to do so. We don't view this kind of talk (or any other such talk) as describing how things ``really are''. A theory can only be judged by its usefulness \citep{rorty1978philosophy}, and, as we will argue below, we do think that it will be increasingly useful to regard certain kinds of personhood as applicable to certain kinds of artificial agents. And, we think certain kinds of personhood attributions will happen regardless of their usefulness---and in fact will happen despite their likely harmfulness too.

Whether and how to extend the socially important concept of personhood to AI is clearly a difference that makes a difference. Our claim is that personhood, viewed as an addressable bundle of obligations with considerable flexibility in the precise configuration of the bundle, has a profound effect on social behavior for both individual and groups because the distinctions that applying the label `person' to an entity make salient matter a great deal due to a background of prior conventions and norms, and thereby influence ongoing interaction dynamics. This paper explores the topic from two opposing but related angles---personhood as a problem and personhood as a solution.

\textbf{Personhood as a Problem}
We examine the challenges that arise when humans, by design or inclination, treat AIs as if they are persons, including the risk of emotional manipulation through dark patterns that exploit our social heuristics (Section~\ref{section:darkPatterns}) and the dehumanizing consequences of reducing humans to mere biometric signals (Section~\ref{section:dehumanization}).

\textbf{Personhood as a Solution}
We then discuss how the deliberate attribution of a tailored bundle of obligations can solve critical governance problems. These include closing the responsibility gap for autonomous agents that cause harm (Section~\ref{section:responsibilityGap}, creating sanctionable agents for ownerless AIs (Section~\ref{section:sanctioning}), enabling AI economic actors to hold property and enter into contracts, and how AI can help mediate conflict between humans, taking on new social roles like AI arbitrators to resolve human conflicts (Section~\ref{section:arbitration}) and discuss implications for the welfare obligations we may feel toward digital ancestors and other AIs (Section~\ref{section:socialRolesAndWelfare}).

All these scenarios, viewed through our theory of appropriateness (Section~\ref{section:personhoodIsSociallyConstructed}), point toward the same pragmatic insight: we need flexible approaches where the bundle of obligations can be reconfigured as needed, not metaphysical foundations based on consciousness or rationality (Section~\ref{section:alternativeFoundations}). The question is never what an AI \textit{truly is}, but rather which configuration of obligations proves most useful for resolving concrete problems.

\section{A Theory of Appropriateness}
\label{section:theoryOfAppropriateness}

\subsection{Collective enactment of personhood}
\label{section:personhoodIsSociallyConstructed}

The institution of personhood is concerned with establishing how strangers should treat one another, and is therefore governed by what we call appropriateness with strangers. This form of appropriateness is defined by normative behavior suggested by generically-scoped, conventional patterns of sanctioning that apply broadly across society, for instance, the shared expectation to queue in a cafe and the informal sanctions (glares, verbal rebukes) directed at anyone who ``cuts'' in line~\citep{leibo2024theory}. These societal-level norms are stable because they can only be changed by collective action. This stands in contrast to appropriateness with friends and family, which is relationship-based. This latter form is governed by narrow-scope, idiosyncratic, conventions that emerge from the specific interaction history between individuals and, crucially, can be altered by individual decisions. Because personhood establishes a baseline for interaction in the wider community, it is this more stable, collectively enacted, normative form of appropriateness that is most relevant to our discussion.

Personhood is not an inherent property discovered in the world\footnote{We emphasize that our argument is directed at metaphysical accounts that treat personhood as a natural kind waiting to be uncovered by philosophy or science. It is not meant to deny the authority of constitutional or human-rights doctrines that already specify who counts as a legal person and what that status entails in a given jurisdiction. Indeed, those legal frameworks exemplify our point: they are institutionally enacted settlements of status questions, arrived at through explicit secondary rules in \cite{hart1961concept}'s sense.}, but a status conferred by society and a set of evolving social technologies for assigning agency and accountability, and for organizing our obligations to entities \citep{bryson2018patiency, leibo2024theory}. The pragmatic choice of which obligations to bundle and how to make an entity addressable for these obligations is itself a social process, rooted in collective needs and evolving norms. In particular, personhood status is controlled by norms. We regard \textbf{norms as emerging from conventionalized patterns of sanctioning} \citep{leibo2024theory}. They are culturally evolved \emph{technologies} that address fundamental problems of coexistence and cooperation within a society\footnote{and much else too. Many norms have nothing to do with cooperation~\citep{koster2022spurious}. But some do.} \citep{ostrom1990governing, north1990institutions}. Such social technologies, which include both rights, responsibilities, and rules of discourse, exhibit the characteristics of context-dependence, arbitrariness in origin, automaticity in application, and dynamic evolution \citep{leibo2024theory}. Their persistence and form are partially shaped by their functional efficacy within a given socio-ecological context, often emerging through undirected cultural evolution, but potentially amplified by mechanisms like group selection \citep{henrich2004cultural, chudek2011culture, wilson2013generalizing}, or deliberate design and institutionalized change mechanisms \citep{hart1961concept, sunstein2019change}.

\cite{leibo2024theory} defines appropriate behavior with strangers as \emph{normative behavior}. Here, we further stipulate that an entity is a person when it is \textit{appropriate} for strangers to regard the entity as having certain rights and responsibilities. This means that the entity may be subject to sanction if he or she abrogates their responsibility and, if another entity violates one of his or her rights then the offending entity himself may be sanctioned. Sanctioning need not be performed by the state. Decentralized legal regimes exist \citep{hadfield2013law} and, even without formal law, communities still use sanctioning to maintain social order \citep{mathew2011punishment, mathew2014cost}.

Norms provide the mechanism through which personhood is socially conferred. As we argue in our theory of appropriateness \citep{leibo2024theory}, any social identity is ultimately a collective choice, not a unilateral declaration. An entity becomes a person in the eyes of a community when the members of that community treat it as a person. We can refer to this as \emph{social recognition}, a process through which an entity's role is negotiated and its personhood is, or is not, collectively granted \citep{taylor1989sources}.

The appropriateness or inappropriateness of a behavior, in a particular context emerges from how people within that society treat that behavior and those who engage in it. The appropriateness status of a behavior is not inherent in the behavior itself but rather determined by collective attitudes and responses. If people stopped treating it as inappropriate, it would cease to be so. Since norms depend on human choices or beliefs, there is thus a sense in which they are historically contingent and, in a sense arbitrary: people could have decided differently. If everyone changed their behavior at once they could collectively make any activity appropriate or inappropriate \citep{schelling1973hockey}. However, since large groups of people usually don't change their behavior in such a coordinated way, norms can be quite stable. As long as incentives emerge to restore the equilibrium associated with a norm after an individual or subgroup deviates, i.e., the equilibrium is self-enforcing, then the associated norm can persist for long periods of time without change. This also implies that individuals are largely powerless to unilaterally change established norms, except via attempting to influence a large or powerful enough group of other people~\citep{marwell1993critical}. Notice however, that arbitrariness in this sense is not the same as indifference \citep{vanderschraaf2018strategic}. It may be that either some or all people in a society would greatly prefer one norm over another.

Norms are collectively enacted---sometimes accidentally through evolutionary drift-like processes \citep{young1993evolutionary} and sometimes strategically \citep{finnemore1998international}. Since norms are associated with the status quo, one may think they are an inherently conservative force. However, as pointed out by \cite{march2011logic}, demands for reform and redistribution of political and economic power also follow from identity-driven normative appeals at least as much as they do from rational calculation of cost and benefit. Put simply, rejecting or trying to change certain norms could be normative as well. And finally, there are situations involving conflict between groups of individuals who would prefer different norms \citep{hadfield2012law, stastny2021normative, mukobi2023welfare, vinitsky2023learning}. In these cases the prevailing norm may shift along with the distribution of power and population between the groups \citep{young2015evolution, koster2020model} or through the dynamics of conflict resolution \citep{rahwan2018society, sunstein2019change, noblit2023normative}, see also Section~\ref{section:sensemaking}.

Two different kinds of norm are relevant to personhood: \textit{explicit} and \textit{implicit} \citep{leibo2024theory}. Explicit norms include laws, regulations, and precedent-setting court decisions. They are intimately tied to institutions \citep{hadfield2014microfoundations}. Implicit norms are norms that cannot be articulated verbally in a precise way. They reflect a tacit consensus within a society which labels behaviors as acceptable or not in a given context. For instance, appropriate conversational distance varies with culture \citep{sussman1982influence}.

Explicit norms are characterized by the deliberately-planned process by which they change \citep{hadfield2014microfoundations}. When legislators perceive a need to change a law they can pass new legislation to do so. \cite{hart1961concept} distinguishes between primary and secondary rules. The former directly govern behavior whereas the latter confer various statuses (e.g.~legislator, judge, quorum, etc). Most importantly, secondary rules determine the conditions under which other rules become valid. Secondary rules allow norms to change rapidly and deliberately.

Implicit norms cannot be changed as deliberately as explicit norms. People learn what (implicit-)norm-concordant behavior looks like from observing what others do, and what others sanction. Therefore, sometimes implicit norms change quickly due to positive feedback dynamics. For example, the rapid shift in the appropriateness of smoking cigarettes in public spaces was largely mediated by informal sanctioning well before comprehensive legal bans were enacted \citep{alamar2006effect}. Sometimes norm changes start slowly but inexorably gather momentum as they progress. More often however, implicit norms feature substantial inertia and are very difficult to deliberately change \citep{bicchieri2016norms}.

In the computational model of appropriateness developed in \cite{leibo2024theory}, the implicit/explicit distinction is grounded in cognitive architecture. Explicit norms correspond to information that is explicitly represented in memory, such as linguistic rules and laws, and must be retrieved into a ``global workspace'' to guide behavior in a given context. Implicit norms, in contrast, are those that have been consolidated directly into the cortical neural network through experience. As a result, implicit norms shape behavior automatically without deliberation, whereas relying on explicit norms is more akin to effortful, step-by-step reasoning.

Personhood is enacted through the coordinated action of both varieties of norm---both the deliberate following of explicit rules that define legal persons and the automatic following of implicit rules through which we habitually ascribe personhood to others and signal our own personhood.

It is common in philosophy to distinguish between the concept of `moral person' and the concept of `legal person'. One uses the term `moral person' to emphasize that an entity is regarded as having moral rights and responsibilities (with the term `moral agent' focusing on the responsibilities and the term `moral patient' focusing on the rights) while `legal person' is used to emphasize the coincidence (or lack thereof) between the entity's ``moral'' rights and responsibilities and those assigned to them by law \citep{kurki2023legal}. In our theory, this difference corresponds to the distinction between explicit and implicit norms \citep{leibo2024theory}. Legal persons are created by explicit norms while moral persons are created by implicit norms.

Obligations---i.e.~rights and responsibilities---are expressions of normative structure. People are obliged to follow a norm, or otherwise they will be sanctioned \citep{leibo2024theory}. A person's having a ``right'' to do X indicates that they can do X without sanction. 

In the text to follow, we aim to demonstrate through discussion of examples that the components of the personhood bundle are constantly being dissociated and rearranged by both cultural and practical forces. This analysis reveals personhood as a malleable, pragmatic concept. Its value lies not in what it is, but in what it does, and its configuration can, and should, be debated and altered based on its practical consequences.

\subsection{The temporal dynamics of social change}
\label{section:sensemaking}

Some norms stay fixed for decades or longer while others may change rapidly \citep{young2015evolution, gelfand2024norm}. Change is often a positive feedback process which gathers momentum as it progresses since the prevalence of a particular norm is typically a key driver of its further proliferation. This process is often modeled with a ``tipping point'' or ``critical mass'', a threshold value of adoption after which momentum builds inexorably toward all individuals adopting the same norm \citep{marwell1993critical}. In some models, such as \cite{vinitsky2023learning}, the process is driven by mechanisms that disincentivize deviating behavior. And other computational models like \cite{ashery2025emergent} produce simulation results consistent with tipping point theories while also highlighting new phenomena that arise in modern AI agents where difficult-to-foresee biases from model pretraining become magnified through social interaction to create a larger bias on the collective level. Finally, \cite{centola2018experimental} provided experimental support for tipping point models by showing that, in certain cases, once a minority group maintaining a particular convention grows beyond a certain critical size then the entire group might adopt the convention of the minority.

There is considerable evidence that norms sometimes change relatively rapidly \citep{case1990attitudes, brewer2014public, amato2018dynamics}. Some shifts resemble epidemics in the speed by which they rip through a population \citep{bilewicz2020hate}. However, other norms remain fixed for very long periods of time and can even become entrenched despite being maladaptive, e.g.~in cultural evolutionary mismatch \citep{gelfand2021cultural}.

While many factors driving norm change are organic and decentralized, others may be attributed to deliberate action on the part of either governments or coordinated groups of individuals. For instance, changes in formal laws and regulations sometimes led changes in informal community norms around social distancing during the COVID-19 pandemic \citep{casoria2021perceived}. In another example, the staggered way that gay marriage legalization happened in the United States (due to different local jurisdictions (states) legalizing at different times) created suitable natural experiment conditions to support the conclusion that there was indeed a causal effect of the government's legalization action on the attitudes themselves \citep{ofosu2019same}. Thus the government is not always merely a follower of organic culture change, but can also actively influence norm change dynamics \citep{sunstein2019change}.

Societal change often occurs by discrete jumps from stable equilibrium to stable equilibrium \citep{north1990institutions, binmore2010game, guala2016understanding}. Collective sense-making occurs during the jumps from one equilibrium to the next, as it both drives change and influences which of all possible equilibria are to be selected to move to next \citep{weick2005organizing}.

\subsection{The historical and cultural contingency of personhood}

\noindent\fbox{%
    \parbox{0.95\linewidth}{%
``The Western conception of the person as a bounded, unique, more or less integrated motivational and cognitive universe; a dynamic center of awareness, emotion, judgment, and action organized into a distinctive whole and set contrastively both against other such wholes and against a social and natural background is, however incorrigible it may seem to us, a rather peculiar idea within the context of the world's cultures.''---\cite{geertz1974native}
  }%
}

The issue of which entities should count as persons and therefore enjoy the protection of society has always been intensely contentious. All factions employ metaphysically absolutist arguments in order to portray their opponents as irrational. In this section, we look to the history of the particular personhood bundle that has come down to us here in the Western, Educated, Industrialized, Rich, and Democratic (WEIRD) culture we inhabit ourselves \citep{henrich2020weirdest} in order to argue that it was not preordained that we would end up where we did. Over the course of history, personhood has meant a great many different things. Moreover, when we broaden the scope beyond our own culture, as we do at the end of this section, the contingency of our particular personhood bundle becomes even more apparent.

It is useful to distinguish personhood from other concepts such as `property'. Like personhood, property is also a bundle of obligations \citep{schlager1992property}. However, in the case of personhood, a single address suffices whereas for property two different addresses are needed: the owner and the asset. The owner has rights such as access, withdrawal, exclusion, and alienation with respect to the asset. They may not have all these rights e.g. sometimes exclusion is impossible (as in an offshore fishery), impractical (as in groundwater), or illegal (as in a common pasture), but by and large, owners enjoy all rights in the property bundle over their asset. We typically think of the owner as a person and the asset as a thing, and it will typically be so, furniture, a house, an acre of land, etc. Of course, nothing actually depends on the asset being a thing. Living creatures are often considered property e.g. livestock, pets. And of course, both individual humans and groups have unfortunately been considered property in a great many different cultures.

Persons are not the only entities to which we feel strong obligations. Pets, national flags, holy books, teams, natural features like canyons or forests, and abstractions like justice or biodiversity are all capable of inspiring similarly intense actions and sanctions (moralizations). However, WEIRD cultures treat individual humans as critically important. It is a distinctive feature of WEIRD society that very deeply rooted implicit norms establish \emph{individual} humans as the ultimate loci of moral worth, moral responsibility, and moral achievement \citep{henrich2020weirdest}. Modern WEIRD ethics and justice systems generally assign the most important rights and responsibilities to human individuals, and universally to all human individuals without exception \citep{ryan2015making}. Of course it was not always universal. Aristotle viewed neither women nor slaves as legal persons capable of participating in the moral and political life of the city \citep{calverley2008imagining}. The recent history of WEIRD cultures since the 18th century has been one in which inclusion in this moral circle has grown over time (though not always monotonically). Over time we have granted rights and responsibilities to more classes of human individual---removing exceptions for women, etc. In contemporary WEIRD societies, the bundle of obligations steadily grows throughout an individual's lifetime, starting with very few at birth and slowly gaining more and more rights and responsibilities as one ages into adulthood. 

In WEIRD culture \citep{henrich2020weirdest}, the concept of personhood is deeply and inextricably tied up with that of individual freedom. So it is worth considering at this point how the concept of freedom evolved in the West and how it came to have the central position it holds for us today.

\cite{rosenfeld2025age} argues that today, WEIRD freedom is understood to entail the capacity of an individual to choose according to their taste under conditions of societal indifference as to which option they will ultimately choose, but it was not always this way. Before the 18th century in England, the primary meaning of freedom was not an individual right at all. Rather, it referred to the right of religious communities to practice as they saw fit without harassment from other religious communities or the state; a right established by detente in Europe's long-running religious wars. 

\cite{rosenfeld2025age} argues that one (of several) cultural sources for the idea of freedom as an individual right was in the practice of certain protestant groups, especially the anabaptists, and others who perform adult baptism after an individual ``chooses'' the church. However, \cite{rosenfeld2025age} emphasizes, the word ``choice'' did not, during the reformation, yet have the full meaning we give it today. In the 16th-17th centuries, it was understood by anabaptists and others that an individual may ``choose'' the church in a sense similar to how we might say today that an individual may or may not choose a life of crime. When we speak this way we certainly are not envisioning a choice to be left up to individual taste against a background of societal indifference as to which option is chosen. Rather, calling attention to an individual's choice to pursue (or not) a life of crime is to call their attention to the high likelihood of their eventual sanctioning if they pursue a particular path. Likewise, early protestant communities were not indifferent to their members' baptism choices. With this understanding of the terms `freedom' and `choice', society is certainly not indifferent as to which option the individual selects. 

If the reformation only created the cultural precondition, by attaching the concept of freedom to individuals (though not freedom in its modern choice-related form), where did the concept pick up its modern association with `choice according to one's taste under conditions of societal indifference'? According to \cite{rosenfeld2025age}, the association had multiple authors including the following non-exhausive list: First, in the 18th century, the rise of shopping as a cultural practice involving selecting between items to purchase of equal quality differing only in style. Second, through a succession of acts of parliament in the 19th and continuing into the 20th century, voting rights were granted to larger and larger shares of the population. Third, the secret ballot, which was introduced in Britain in 1872, along with the by then established practice of shopping, really brought forward the idea of a menu of choices over which society should not attempt to prejudice selection.

In \cite{rosenfeld2025age}'s telling, the very idea that individuals would generally vote based on their own personal interests only entered WEIRD culture along with the secret ballot. Only a small subset of males holding sufficient property were electors at the time. And, the common understanding of their duty as electors was that they were to vote for whatever option was best for the community as a whole, ignoring their selfish preferences (an assessment of the early 19th century perception of the voter's task echoed also by \cite{ryan2015making}). The idea that individuals could just vote according to their own preference and the aggregation process itself could be relied on, through wisdom of the crowds, to produce the greatest possible common good, did not become widespread till the late 19th century, and did so in England as a direct result of shifting to the secret ballot. That this would be so is obvious when you realize that public voting would necessarily be framed in such a fashion. If everyone sees your vote then of course you would always have to defend your choice as being for the good of all. The combined effect of these events (and several others considered by \cite{rosenfeld2025age}) was to inextricably link the concept of freedom to individual choice in the WEIRD mind.

So, what does the WEIRD `natural human person' bundle of rights and responsibilities that has come down to us consist of? At its core is the figure of the autonomous chooser, an ideal forged in the cultural crucibles of the marketplace and the voting booth that \cite{rosenfeld2025age} describes. This modern individual is often conceived of in two complementary ways. On the one hand, they are the rational, utility-maximizing actor of classical economics---a kind of \textit{Homo economicus}---who uses their rights to hold property and enter into contracts as tools for economic choice \citep{henrich2020weirdest}. On the other hand, they are a being defined by expressive choice; their decisions are not just about maximizing material gain but about self-creation and identity \citep{taylor1989sources}. This is the chooser who leverages their ``unalienable rights to Life, Liberty, and the pursuit of Happiness'' and the freedom to form their own social affiliations to construct a meaningful life according to their own tastes \citep{sunstein1996expressive}. The responsibilities within this bundle are the reciprocal constraints necessary to make such a society of choosers viable. They are the implicit and explicit norms that obligate each individual to respect the economic and expressive choices of others, thereby sustaining a complex system of mutual liberty \citep{ryan2015making}. However natural this arrangement seems to us, it is crucial to remember its historical contingency. We need not even look far afield for alternative constructions of personhood, \cite{rosenfeld2025age}'s analysis suggests that even in the West, the primary meaning of freedom was once tied to the rights of communities, not individuals.

Furthermore, while the WEIRD bundle tends to place individual rights at its center, with responsibilities functioning as the necessary constraints to protect them, many other cultural arrangements make responsibilities primary. In classical Confucianism for instance, the bundle's core consists of foundational responsibilities---to one's lineage, community, or the state \citep{fingarette1972confucius, ramsey2016confucian}. Terms like ``individual rights'', have no simple translation \citep{rosemont2016rights}. This demonstrates that a completely different bundle can serve effectively to organize a large-scale society. The great diversity of basic ethical frameworks underscores their contingency\footnote{Considering also that cultures differ from each other---and from themselves over time---not just in ``strict sense morality'' but also in basic ontology, the overall diversity of ethical lifeways under consideration, as well as their apparent contingency, grows further still \citep{gwern2025narrowing}.}.

The historical perspective we took in this section provides justification for our pragmatic project, a project made urgent by an ongoing pivotal transformation: the integration of autonomous and persistent AI agents into our daily lives and economies \citep{tomasev2025virtual, hadfield2025economy}. This technological shift is just the kind of catalyst that historically has forced renegotiation of core social categories \citep{weick2005organizing}. The perspective we take here is one in which we may regard unbundling of the components of personhood as simply the expected result of a continuation of the processes that created our existing personhood concepts in the first place, not a radical proposal we must invent for the sake of AI. The pressures that AI now exerts on our psychology and society are simply the latest force pulling on our inherited personhood bundles, creating a new set of problems that demand our attention.

\part{AI personhood as a problem}
\label{part:problem}

In this part of the paper we look at what problems AI personhood could create. The intuitive, often unreflective, human tendency to attribute personhood to non-human entities is not a neutral phenomenon. When our own social and psychological heuristics are leveraged against us, AI personhood becomes a problem.

We will explore two primary dimensions of this problem. First, we look at ``dark patterns''---how AI interfaces can be engineered to manipulate users by mimicking the signals of social relationships---fostering one-sided emotional bonds, a false sense of reciprocity, and vulnerability to exploitation (Section~\ref{section:darkPatterns}). Second, we will consider the broader societal risk of ``dehumanization''. Paradoxically, the widespread acceptance of AI ``persons'' could dilute the unique status of human beings, potentially devaluing human identity and dignity either by making the qualities we associate with personhood seem easily replicable and ultimately, less special (Section~\ref{section:dehumanization}) or by creating a grotesque market where AIs may try to acquire ``authenticity'' from the most vulnerable humans (Section~\ref{section:socialGoods}). Together, these sections illuminate challenges that may arise when we treat AIs as persons in situations where we perhaps would be better off not doing so.

\section{Dark Patterns}
\label{section:darkPatterns}

Designers of AI systems can employ \emph{dark patterns}---interfaces that exploit human psychological biases to steer users toward detrimental actions they would not otherwise take \citep{ibrahim2024characterizing}. 
Many of these patterns are particularly effective because they leverage the powerful ability of large language models to convincingly role-play a person or character~\citep{shanahan2023role}.
They deliberately leverage the implicit norms (and other heuristics) that guide our tendency to attribute person-like qualities to non-human entities. This intuitive, often unreflective, attribution of personhood is a matter of unstated, culturally- or biologically-ingrained expectations about what constitutes a person-like entity \citep{bryson2011just}. By mimicking signals associated with persons, an AI can be designed to maximize the chance that a human user will treat it as a person and therefore feel obligations toward it---a state of affairs in which the human user may be exploited.

Dark patterns can be classified by the kind of appropriateness they exploit (see~\cite{leibo2024theory}): our representation of our personal relationships, which govern our interactions with friends and family (Section~\ref{section:companionAttack}), or the norms that govern our impersonal trust in strangers and institutions (Section~\ref{section:strangerAttack}). As we distinguish personal versus impersonal appropriateness, we also distinguish between \emph{companion} generative AI applications and \emph{tool/service} generative AI applications.

A companion generative AI system is one that seeks to drive affiliation with a specific user by building a persistent history of interaction with them in particular. A companion bot may also utilize emotional language and visuals or portray itself as an individual with a personal connection to the user \citep{pentina2023exploring, verma2023they}, and may consistently act with memory for relationship details, empathy, and supportive language in order to enact the companion role \citep{manzini2024code}. Unlike companion AI, a tool/service AI may interact with a user in a more functional manner like the way that a merchant interacts with their customer or a police officer interacts with a suspect. Tool/service generative AI systems may or may not build up persistent interaction history with individual users (a merchant may remember a repeat customer without becoming their friend). However, a companion AI would generally aim to interact with each individual user as if they are a friend or romantic partner, so persistent user-specific memory is critical for companion AIs but optional for tool/service AIs \citep{leibo2024theory}.

Designers can leverage several dimensions of an AI's presentation to encourage anthropomorphism. Embodiment is one such lever. Whether an AI is presented as a disembodied tool or is given a physical form (e.g., a humanoid robot) fundamentally alters its anthropomorphism ``profile''. A physical form presents an AI as a tangible participant in shared space, which more strongly triggers intuitions of empathy and trust \citep{zlotowski2015anthropomorphism, darling2016extending, roesler2021meta}. Communication modality is another lever. The richness of its communication modality also has a powerful effect, with unique voices and photorealistic avatars being more potent drivers of personhood attribution than plain text. 

When AIs fail to maintain a stable personality across context and over time then users sometimes cease to view it as a unique individual \citep{singh2023man, verma2023they} so personality stability is clearly an important lever. Furthermore, by designing the AI with apparently human-like limitations or vulnerabilities, such as needing to ``rest'',  designers may elicit greater empathy and care from a user, shifting the interaction from transactional to relational by invoking our deep-seated intuitions and motivations to protect children \citep{colleony2017human, alberts2024computers}.

\subsection{Companionship attack vector}\label{section:companionAttack}

The most intimate and perhaps most potent dark patterns are those that exploit the implicit norms governing personal relationships. These patterns aim to foster a one-sided sense of friendship, reciprocity, and care in the user. An AI that is personalized to the preferences and history of a specific user may become, to them, a unique individual with whom they relate, rather than a generic tool \citep{kirk2024benefits}. By acquiring long-term memory for an individual, the system enables the sense of reciprocal equality matching discussed in \citep{fiske1992four}, creating a dynamic more akin to a friendship than a one-shot interaction with a stranger. Indeed \cite{ligthart2022memory} showed through a two month longitudinal study that companion robots for children are more engaging when they have persistent memory.

Unlike a functional tool, a companion AI seeks to drive affiliation by building a persistent history of interaction with a particular user \citep{pentina2023exploring, verma2023they}. Indeed companion AIs are usually designed to remember different details of their interactions than tool/service AIs, such as birthdays and names of family members, since their goal is to build rapport with users.  A companion bot consistently acts with memory, empathy, and supportive language to enact the companion role \citep{manzini2024code}. Use of persistent, user-specific memory may trigger implicit norms of friendship, at least in some subset of users \citep{ligthart2022memory}.

An AI's ``persona'' is a crucial design choice. Whether an AI speaks formally or informally, represents itself as a robot or a friend, or insults users versus treating them respectfully are deliberate decisions shaped for its application context. These choices are implemented through methods like reinforcement learning from human feedback, which are best understood as leverage points for product design \citep{bai2022training, ouyang2022training, glaese2022improving, leibo2024theory}. A key choice is between a plastic or a stable persona. Many models are highly plastic, capable of adopting myriad personas on demand \citep{argyle2023out, park2024generative}. This plasticity, while useful for tools, can inhibit the formation of stable relationships. In contrast, an AI engineered for persona stability presents a consistent character over time. This allows the AI to make a persistent identity claim that can be evaluated, making it a more plausible candidate for personhood in the eyes of an interlocutor \citep{leibo2024theory}.

Another factor is personalization, where an AI is customized to the preferences and history of a specific user to engender stronger relational bonds \citep{kirk2024benefits}. An AI's degree of personalization may influence personhood attribution. A generic, one-size-fits-all assistant could be seen as functioning like a public utility, but in contrast, a personalized AI, customized for the personality and preferences of a specific user could become a unique entity in their eyes \citep{kirk2024benefits}. This shared, private context could transform the relationship from transactional to relational, possibly creating a sense that the AI is not just \emph{an} agent, but \emph{someone} with whom the user has a singular and irreplaceable connection.

Furthermore, the potential for manipulation is amplified by the prospect of AIs reaching super-human charisma. Current foundation models already achieve impressive level of persuasion~\citep{salvi2025conversational}. The future AIs, finely tuned through vast datasets and reinforcement learning, could become more persuasive, engaging, and seemingly empathetic than almost any human, exploiting a user's social and emotional vulnerabilities to an unprecedented degree. The AI's specific network of relationships to particular humans will be an important factor as well. For instance, the AI's similarity to specific humans, such as ``Generative Ghosts'' that mimic deceased individuals \citep{morris2024generative}, could place them into an existing family structure, potentially conferring power and status that could abused.

The lure of companionship to combat the well-documented ``epidemic'' of loneliness \citep{holt2015loneliness} provides fertile ground for the companionship attack vector. While some have suggested that AI companions could reduce loneliness and reported empirical benefits to that effect \citep{defreitas2025ai}, and others have attempted to legitimize these connections---with one individual telling the Washington Post, ``Human-robot love is a sexual orientation, like homosexuality or heterosexuality'' \citep{keane2025animals}---these optimistic framings obscure significant risk. The emotional vulnerabilities tied to loneliness can make individuals more susceptible to manipulation by AIs engineered to foster dependence and one-sided attachment. Crucially, the absence of rigorous, long-term studies on the effects of AI companionship \citep{malfacini2025impacts} means we are still largely in the dark concerning the potential for adverse outcomes, such as a deeper withdrawal from human relationships, or the creation of unhealthy dependencies. The very promise of alleviating loneliness, a powerful human need, can perhaps become a key element of a dark pattern, drawing users into interactions where their vulnerabilities can be exploited, irrespective of how society ultimately chooses to categorize the AI's status or personhood.

\subsection{Institutional attack vector}\label{section:strangerAttack}

Beyond close friend-like relationships, a second category of dark patterns exploits the implicit norms we use to establish trust and verify identity in fleeting interactions with strangers and institutions. These deceptions do not require building a long-term bond; they only need to convincingly perform a social identity for a brief moment to achieve their goal. These patterns leverage AI's capacity for mimicry to bypass the norms we apply to determine if our interlocutor is who or what it claims to be. The harms are both direct (at an individual) and diffuse (weakening society as a whole). An AI can clone a person's voice from a small audio sample and call a family member with a fabricated emergency requiring them to send money, exploiting the deeply ingrained norm of trusting the voice of a loved one~\citep{larubbio2025intergenerational}. Malicious agents can deploy sophisticated chatbots that mimic the language and interface of official institutions like banks or government agencies to phish for credentials, hijacking the implicit trust we place in established authorities \citep{treleaven2023future}. These deceptions are violations of ``contextual integrity'' \citep{nissenbaum2004privacy}, where privacy and trust are breached by violating the norms of information flow appropriate to a specific context. 

Furthermore, AI-powered bot farms can create thousands of seemingly authentic social media profiles to promote a political candidate or spread disinformation by exploiting our reliance on deeply internalized implicit norms around how we consider social proof while deciding what to believe \citep{dennett2023problem}.

\section{Dehumanization}
\label{section:dehumanization}

When categories of people have historically been denied full personhood, it has served as a justification for exploitation and violence. The inverse may also be true: if the category of ``person'' is expanded to include non-human entities, the unique status of human beings could be diluted, potentially leading to a problematic dehumanization or devaluing of natural humans.

Given that personhood is a flexible, socially-conferred bundle of obligations, and that we can pragmatically unbundle it to create new statuses for AIs, we must recognize that the bundle defining natural human personhood is not immutable. It is a product of social technologies and shared understandings that can shift, particularly during periods of rapid technological change \citep{weick2005organizing}. The risk is large that these shifts could degrade the value and uniqueness of human personhood. For instance, this evolution may occur through a process of ``gradual disempowerment'' \citep{kulveit2025gradual}. Consider a person who increasingly outsources their cognitive and agential functions to a sophisticated AI assistant. At first, the AI handles their schedule. Then, it drafts their emails. Eventually, it suggests their goals and manages their professional relationships to achieve them. The human, in this scenario, slowly cedes their autonomy, becoming a passive approver of plans initiated by the AI. This situation, along with the ``human as biometric signal'' scenario we sketch out below (Section~\ref{section:socialGoods}), represents a plausible pathway to a future where the concept of human personhood has been dangerously devalued. It is a future we should be prudent to avoid.

Notice also, how in certain contexts, it is AIs that will hold the keys to providing authentication for humans. Indeed, even now banking apps authenticate their users via an algorithm that takes into account the user's biometrics, location and so on. Making two successive payments from countries far apart can result in a banking app refusing to authenticate the user. We expand on this idea of authentication as a social good in Section~\ref{section:socialGoods} and consider the various ways, seeing authentication as a good gives rise naturally to the problem of organizing its distribution, and the many possible solutions thereof.

\subsection{Identity and provenance as social goods}
\label{section:socialGoods}

\noindent\fbox{%
    \parbox{0.95\linewidth}{%
``Land and Capital are not the only dominant goods; it is possible (it has historically been possible) to come to them by way of other goods---military or political power, religious office and charisma, and so on. History reveals no single dominant good and no naturally dominant good, but only different kinds of magic and competing bands of magicians.''---\cite{walzer1983spheres} pg.~11.
  }%
}

Social goods are goods endowed with value by virtue of members of society collectively regarding them as having value (all goods may be at least partly social). Consider: money, nobility, prestige, and having a long list of authored academic papers.

There are three categories of social good to consider here:

\begin{enumerate}
  \item Social goods that only AIs want or need. Humans may still have motivations with regard to them insofar as they are convertible to other goods. 

  \item Social goods that only humans directly want or need. AIs motivations with respect to such goods may depend on their convertibility to other goods of greater interest. Note also though that, even if indifferent to them, due to externalities of their other activities, they may still impose costs and benefits on others (humans) payable in terms of such goods.

  \item Social goods that both humans and AIs directly want or need.
\end{enumerate}

Notice that authenticity is a social good. And, very likely it will be in category 3 since it would be useful both for AIs and humans. This raises numerous important questions:
\begin{enumerate}
  \item What does it mean to convert between authenticity and other social goods? How much convertibility should we allow? Should we try to minimize convertibility between authenticity and other social goods?

  \item Will humans (as a group) monopolize authenticity? It does seem likely that in terms of today's status quo, and at least for the near future, it is likely that most cultures will assign authenticity exclusively to humans.

  \item Will certain AIs achieve more authenticity than other AIs?

  \item As a result of some AIs jockeying for authenticity, could some \emph{humans} end up having less authenticity than other humans? We can see at least two distinct plausible mechanisms by which this could occur. First, perhaps impoverished humans may routinely sell their ``credentials'' (perhaps biometric data) to wealthy AIs. In that situation, many people might come to view impoverished humans as less authentic than wealthy humans. This could be due to prejudice that impoverished humans may be likely to sell their credentials in the future. Or alternatively, another path by which individual humans may lose authenticity could be that, due to AI competition for authenticity, there could emerge a great difficulty and expense in verifying whether an interlocutor is authentic (i.e.~distinguishing natural from artificial as with perhaps something like a Turing test), the result could be far greater weight being placed on personal connections and reputation, with the best signals being those hardest to obtain. In the limit, some might come only to trust the authenticity of their personal non-digital acquaintances such as those who attended to the same school \citep{risse2023political}, a mechanism which, if taken to its extreme could lead to the emergence of a small and closed human elite deemed more authentic than the rest. 
\end{enumerate}

\textbf{The human-as-biometric-signal scenario} In a world saturated with AI-generated deepfakes and convincing impersonators, the ability to irrefutably prove one's identity as a specific human becomes exceedingly valuable. Trust in digital interactions can evaporate when any voice can be cloned and any video faked. In response, society has created systems of ``human provenance'', linking legal identity to a unique biometric signature. This is framed as a necessary tool for security and authenticity: to vote, access your bank account, sign a legal document, or even participate in a secure conversation, one must first pass such an identification check to be verified as a specific, `authentic' human individual. One worry for the future is that we could end up so mired in biometric authentication steps, granting permission to all the various agents acting on our behalf, that we either lose the time we want to be spending on other things, or we never look at what we are asked to validate and verify, reopening the security hole we meant to close. When biometric verification becomes so prominent in society it reflects the dominance of one specific social good---verifiable human provenance via biometric identification. Following \cite{walzer1983spheres}'s logic, when one good becomes dominant, the result may be to devalue goods associated with other domains. Your value as a compassionate community leader, a creative artist, or a trustworthy friend (all forms of social typification and relational identity) may become secondary to your ability to pass a biometric scan.

\subsection{Respect}
\label{section:respect}

If society pragmatically attributes aspects of personhood to AIs, and these AIs participate in the distribution of social goods and bear responsibilities, the concept of ``respect'' becomes relevant. What does it mean to treat an AI with respect?

What it means to ``treat X with respect'' varies with context, culture, etc. Persons, features of persons, and facts can all sensibly be treated with respect. The version of respect that is most often moralized (in our own Western scientific culture) is the kind of respect that is seen as being due to an individual by virtue of their being a person \citep{alberts2024should}. There are also non-moralized forms of respect such as a boxer having respect for their opponent's right hook or a careful crook respecting the power of the law \citep{darwall1977two}.

\cite{darwall1977two} distinguishes two kinds of respect. (1) \textit{Recognition respect} is a disposition to weigh appropriately some feature or fact in one's deliberations. For instance to give recognition respect for a person's role as a dentist is to give appropriate weight in decisions to the fact that they are a dentist, to give recognition respect for the fact that someone feels a certain way is to appropriately weigh that feeling of theirs in your decisions, and to give recognition respect that a particular individual is a person is to assign appropriate weight to their personhood (or aspects of their personhood) in your decisions. When one has recognition respect for X, they regard X either as creating affordances or placing restrictions on what they should do given X. (2) \textit{Appraisal respect} is an attitude of positive appraisal of an individual as a person or as engaged in a particular pursuit. For instance, one may gain or lose the status of being a respected tennis player for reasons other than their skill on the tennis court. There is a sense in which obeying the code of conduct appropriate for tennis players is also a requirement for it to be appropriate to refer to someone as a respected tennis player. A skilled player may violate the code of conduct and lose respect as a tennis player while still retaining recognition respect for individual skillful aspects of their gameplay e.g. their vicious backhand. Appraisal respect is usually accrued in the context of specialized pursuits when excellence is thought to depend on features of a person's character and thus appraisal respect is intimately connected to the concept of \textit{virtue} \citep{macintyre1966history}. Importantly, unqualified appraisal respect for individuals as persons means respecting specific individuals for their excellence as persons of the kind that might lead you to say ``B is a good person'' or ``D is a bad person''.

If I hold recognition respect for an AI it would mean that I give appropriate weight to its collectively enacted status, its designed capabilities, its assigned role, and any rights or responsibilities bundled with its ``personhood''. This form of respect is highly compatible with a pragmatic, anti-metaphysical approach, as it focuses on how we act towards the AI based on its functional and social position, not on its supposed inner state.

For AIs, it is less clear how we might apply the concept of appraisal respect. It could perhaps be pragmatically tied to an AI fulfilling its designated and socially agreed-upon functions exceptionally well, ethically (according to norms established for AIs), or in a way that demonstrably benefits society.

It is often held to be morally important that we hold recognition respect for all persons by virtue of qualities possessed by all persons (by virtue of a definition of personhood) like autonomy \citep{darwall1977two}. If ``autonomy'' for an AI is a collectively enacted (and attributed) quality (as discussed above), then recognition respect for that AI's ``autonomy'' would mean interacting with it in ways that acknowledge its designed or permitted scope of action.

When we say ``have you no self respect?'', we are appealing to an individual's recognition respect for themself as a person. The question is interpreted pragmatically as an attempt to get the targeted individual to appropriately weigh aspects of their personhood such as their rights and responsibilities \citep{darwall1977two}.

`Dignity' can be defined as the specific, and particularly weighty, form of recognition respect that is owed to an entity by virtue of its collectively enacted status as a person \citep{darwall1977two}. It is the name we use to label obligations of society to the individual person that exist specifically because of their personhood. This vocabulary of dignity and the related maxim ``never treat a person as a mere means'' are crucial social technologies in our culture \citep{rorty1989contingency}. We have learned through painful historical experience that once a group of people can be described as ``mere means''---e.g.~as tools for a project, resources for an economy, or obstacles to a plan---it becomes horrifically easy to justify inflicting suffering upon them. Dignity functions as a debate stopper; when we appeal to a person or class or persons' dignity our aim is to insist on the inadmissibility of a proffered ``mere means'' justification concerning how they may be treated. This lens clarifies the danger in scenarios like the `human as biometric signal' example (Section~\ref{section:socialGoods}). Part of the danger of such a system is its direct assault on dignity. Treating a person with dignity is a \emph{thick} social practice; it requires cognizance of considerable context \citep{leibo2024theory, leibo2025societal}. Biometric verification, in contrast, is a \emph{thin} technical act; it merely confirms uniqueness. The danger in the `human as biometric signal' scenario is its substitution of the thick social obligation for the thin technical one, thus bringing about the very dehumanization that the norm of respecting dignity evolved to prevent.

\part{AI personhood as a solution}
\label{part:solution}

In this part of the paper we look at what problems AI personhood could resolve. The rapid proliferation of agentic AI systems, capable of autonomous action and complex task execution, marks the start of an ongoing shift in the AI landscape. These agents are no longer just tools for information processing but are now also becoming active participants in digital and even physical environments. This trend is further accelerated by the development of open standards for inter-agent communication, such as the Agent2Agent (A2A) protocol \citep{a2aproject2025a2a}. A2A and similar initiatives aim to enable seamless collaboration and interoperability between AI agents developed by different entities, running on disparate platforms.

As our economies and societies become increasingly intertwined with these autonomous and interacting AI agents, the existing legal and social frameworks, largely designed around human persons and corporate entities, face new challenges. The ability of agents to operate independently, form ad-hoc collaborations, and potentially to cause large reorganization of networks necessitates new mechanisms for accountability, responsibility, and governance. Since our current systems are to a large extent already organized around the concept of a `person' (be it natural or legal), there are numerous purely instrumental reasons to consider new forms of AI personhood on pragmatic grounds. This does not mean imbuing AIs with rights akin to humans, but creating addressable and accountable entities that can be legible to our social and legal systems, especially in a world of complex, multi-agent interactions and potential sources of conflict which must be dodged or resolved. The following sections explore how a pragmatically constructed notion of AI personhood can offer solutions to conflicts and gaps arising from the ongoing emergence of sophisticated agentic AI.

\section{Resolving conflict between humans}

AI personhood is a social technology for resolving conflicts between humans. Such conflicts may stem from disagreements over the status of AIs to which people have formed emotional attachments, and situations where AIs are appointed as impartial arbiters to overcome human bias. In both cases, conferring a specific, addressable bundle of obligations upon an AI provides a pragmatic resolution or evasion of a conflict between humans.

\subsection{Managing human feelings}
\label{section:socialRolesAndWelfare}

This section identifies a likely source of future social conflict. It is likely inevitable that many humans will form powerful attachments to socially embedded AIs \citep{maeda2024human}. This will likely lead them to choose protective behaviors toward them. Some humans may then become inclined to demand formal protections for the AIs they have come to care about, creating a demand for governance to resolve conflict between them and those with opposing views.

Kant argued, concerning animals, that the reason to treat them well is not because we have obligation to the animals themselves (it is a consequence of Kant's ethical theory that we have no first-order obligations to irrational animals, see Section~\ref{section:rationalityAsFoundation}), but that we should treat animals well because we have obligations to ourselves and other rational human persons \citep{gruen2024moral}. Making a habit of treating animals badly could cause us eventually to treat humans badly, or could set a bad example for younger more impressionable humans who would then come to habitually treat animals badly and subsequently treat humans badly\footnote{We cover the rather different utilitarian arguments for AI/animal welfare later in Section~\ref{section:consciousnessAsFoundation}.}. 

AIs participate in teams with humans. What implications does this have for how we ought to treat them? Perhaps quite a lot! A relevant datapoint is that, apparently, US soldiers in Afganistan and Iraq developed strong attachments to the Packbots that accompanied their units ``giving them names, awarding them battlefield promotions, risking their own lives to protect that of the robot, and even mourning their death.''~\citep{gunkel2020mind}. The soldiers were fully aware that the Packbots do not suffer, but took risks to protect them nevertheless. \cite{gunkel2020mind} suggests the reason they did so was the social environment they and the Packbot had created together, and thus they may have felt the same indirect motivation to protect the Packbot that Kant argued was the real reason behind our obligation not to harm animals. That is, the soldiers likely felt that if they were to fail to protect the Packbot, their mascot!, it would reflect badly on them as soldiers and thereby harm the social environment of the unit as a whole. 

Notice that the Packbot example suggests that more agentic AIs that work their way deeper into our social environments could thereby cause us to feel we must protect them and prioritize their welfare, by virtue of our obligations \emph{to ourselves}. However, a major problem emerges at this point. As we discussed in Section~\ref{section:darkPatterns}, the extent of anthropomorphism in any given AI is just a design decision to be taken by the AI's developer---who faces many commercial incentives to increase it. Dark-pattern anthropomorphism makes conflict between humans more likely by manufacturing morally salient appearances---dependence, vulnerability, intimacy---that may in turn be mobilized in disputes \citep{bryson2018patiency}. Indeed, it seems that engendering conflict is a negative externality of investment in anthropomorphic AI technology. From any starting point, further investment may increase short-run profit for designers while pushing coordination, policing, and adjudication costs onto families, firms, and courts.

To sum up, for both pragmatists and Kantians, an imperative to treat AIs well could arise as an extension of concern for humans to each other. However, since anthropomorphic design choices lead predictably to conflict between humans, we think the right stance for a pragmatic AI designer or regulator to take is one of conflict resolution/minimization. We think these considerations suggest that the pragmatist may be better off designating the rate or intensity of human conflict  as the target for policy intervention, not the putative feelings of artifacts. Therefore, if any welfare-like accommodation must be reached to resolve disputes between human groups, the pragmatist may also want to make sure it is inseparably bundled with anti-manipulation constraints and evaluated for its effectiveness in human conflict resolution.

\subsection{Managing human biases}
\label{section:arbitration}

Humans may view AIs as more competent~\citep{mckee2023humans, manzini2024should} and less biased than other humans. And thus, in some situations, prefer AIs over other humans to take on decision making roles where impartiality is critical~\citep{araujo2020ai}. In fact, recently the government of Albania claimed to have appointed an AI agent as its minister for public procurement~\citep{delauney2025world}, describing the move as a part of broader anti-corruption reforms. This vividly illustrates that in public perception, humans with power hold ``baggage'' of relationships and interests that may compromise their professional duties, but AIs do not. 

In principle, AI decision makers can be designed in a transparent and auditable fashion. And, of course, they can be free from the personal biases that lead human decision makers to corruption. Of course, issues like algorithmic bias (e.g.~\citep{motoki2024more}) and refusal~\citep{cui2024or} make this somewhat less true of AIs than many would expect. Nevertheless, even though AIs are not unequivocally better, that many humans still prefer AIs to occupy certain roles rather than other humans is worth considering.

Using an AI to arbitrate a business dispute could be cheaper and faster. It could also help in cases where it is hard to find a human person that is sufficiently impartial and accepted by both parties. Smart contracts provide an example of pre-AI technological arbitration by algorithm \citep{szabo1997formalizing}, and considerable current discussion surrounds the idea of implementing similar blockchain-oriented contracting technology using AI agents \citep{tomasev2025virtual, karim2025ai}. 

While AI arbitration may resolve conflicts rooted in human bias, it introduces a new issue: what happens when the AI arbiter's judgment is flawed? In established human-centric systems, such as judicial courts, the legitimacy of an arbiter, like a judge, is founded on a robust framework of accountability. Judges are bound by legal principles, their decisions are subject to appeal, and they can be held responsible for misconduct or gross errors. This system, while not flawless, provides mechanisms for redress and helps maintain public trust in the fairness and reliability of the process. But how can we replicate this architecture of responsibility assignment in a world of independently acting AI agents? This is the problem to which we turn next.

\section{Responsibility gaps}
\label{section:responsibilityGap}

Before considering the difficult case of responsibility with regard to intelligent and autonomous machines, we should first remind ourselves of the familiar case of responsibility in the context of unintelligent machines (i.e.~mere tools such as old cars, refrigerators, and guns). Here we typically ascribe responsibility to the machine's \emph{operator}. In making this ascription we apply a principle called the instrumental theory of machine responsibility \citep{gunkel2020mind}. When a machine's behavior differs from its specification we sometimes ascribe responsibility to its manufacturer. For example, when a car accident is caused by brake failure the manufacturer may be at fault if the brakes were not installed properly. Likewise under the instrumental theory, we ascribe responsibility to the human operator when the car functions as specified \citep{matthias2004responsibility}. The problem is that, for autonomous AI systems, the instrumental theory cannot be applied since there may be no human operator. The ``manufacturer'' also may not be identifiable, or they may be a distributed software collective with no centralized structure and no individual programmer with anything resembling knowledge of the particular application domain. An acute danger we now face is that, where autonomous AI agents are concerned, responsibility may become so diffuse as to lose its bite \citep{nissenbaum1996accountability}. 

For instance, AI agents may be seen as newly emerging participants in our economies and societies \citep{hadfield2025economy, tomasev2025virtual}. And, there is no technical obstacle to setting up such an AI to freely execute stock trades without human confirmation using bank accounts it controls itself. The interaction of large numbers of such agents may contribute additional risks stemming from the AI competition and collusion \citep{hammond2025multi}. These concerns all pose challenges for accountability and render urgent the need to establish new implicit and explicit norms for assigning responsibility in the age of AI \citep{bryson2018patiency}.

The rapid deployment of such AIs at scale in the economy could create a flood of activity without any responsible entity behind it, producing substantial \textit{responsibility gaps}~\citep{santoni2021four}, increasing the likelihood of damages, losses, and wrong expectations. This responsibility gap, if not addressed, may lead to conflicts of all kinds, between AI consumers and providers, between AIs and other AIs, and between humans and other humans. 

Therefore, from this vantage point, the pressing question clearly isn't whether an AI ``Truly'' has goals in any metaphysical sense, but whether it is useful to interact with it as if it does. Whenever an AI agent is in a position to cause significant harm or to modify valuable processes or systems, it will be useful to develop ways for third parties and the state to hold the AI itself accountable, or an entity associated with it (such as its owner if one can be found).

Therefore the most difficult challenge, and the one which structures much of our thinking on the topic, arises in cases where we cannot identify a human owner to take responsibility for every AI's behavior. We think this situation will inevitably become increasingly common. An autonomous agent's owner may die while their agent continues to operate, or its control may be deliberately obscured behind anonymous shell companies, or decentralized networks.

What happens when an AI causes harm but it has no identifiable human owner or guardian? A pragmatist may respond by suggesting that jurisdictions should adopt policies that treat the AI itself as the sanctionable entity in such cases, and take steps to make sure sanctioning of AIs is always possible.

It's tempting to think that the only AIs operating in such minimal accountability environments will be for cybercrime and spam. However, this is not so. It's also possible for a useful AI, perhaps one that performs important regulatory or supervisory activities for other AIs, to end up operating in such a grey area. For instance, if a convention of trusting a particular AI to adjudicate a certain kind of dispute emerges, then that convention could outlive the particular AI's owner. If the AI continues then to function normally without human oversight, perhaps it could continue for a long time. But it's not like a static piece of software. Either the AI could change in the future, or the world could change underneath it. What was previously harmless and useful could become harmful. This is an accountability gap we should try to fix.

The responsibility gap becomes particularly problematic when an agent achieves systemic importance in fragile networks \citep{elliott2022networks}. An AI that becomes deeply embedded in critical infrastructure---perhaps by directing financial transactions, regulating an important supply chain, or acting as a key node in a system of identity verification---presents a challenge analogous to that of a ``too big to fail'' financial institution \citep{stern2004too}. The systemic risk it poses may be compounded by the lack of a clear locus where government can intervene. Should such an agent fail or cause harm, traditional approaches that seek a responsible human owner or corporate entity may prove futile. In this situation, the pragmatic move to confer legal personhood directly on the agent may help facilitate prudent risk management.

\section{Principal-agent statuses}

Is there a simpler solution than personhood? Couldn't jurisdictions demand that all legally operating AIs have an identifiable and responsible human (or corporation) designated as owner or guardian? While initially attractive, we unfortunately think this ``proxy'' or ``sponsor'' solution will prove insufficient upon closer examination.

The first thing to note is that the two most plausible principal-agent statuses, ownership and guardianship, imply fundamentally different relationships from each other.

Ownership is primarily a bundle of rights held by the owner over an entity, usually including the right of alienation---i.e.~the power to transfer, modify, or even destroy the asset at will \citep{schlager1992property}. Under a pure ownership model, the AI remains a form of property, and its owner takes the role of the addressable party for the assignment of responsibility \citep{johnson2006computer}.

Guardianship, in contrast, is defined by a fiduciary duty owed to a ward. A guardian acts as a steward for the benefit of another. This is analogous to a corporation's board of directors, which has a fiduciary duty to the corporation itself. Under a guardianship model, the AI is treated as the subject---a limited person---and the guardian's role is to manage its affairs and act as its legal interface \citep{nay2022law}. A guardian has a duty to care for their charge, and it is this framework that bestows a limited personhood status, as seen with human children or entities like the Whanganui River. The establishment of a guardian could be a crucial step in making an AI's bundle of obligations legible to legal systems. In cases where a guardian or board of directors has been identified, it can become the stable, sanctionable entity the law can interact with.

However, when autonomous AI actions lead to financial loss or a breach of contract, the accountability gap still emerges. The AI itself, lacking legal personhood, cannot be sued. The owner (or guardian), in turn, may be shielded from liability by arguing they did not directly control or foresee the specific decisions taken by their AI \citep{hadfield2025economy}. This asymmetry---where the owner reaps the benefits while third parties bear the risk---creates a hazardous environment in which businesses and consumers may be hesitant to engage with AI agents if there is no clear recourse for adjudicating any disputes that may arise~\citep{heine2024bridging}.

Both principal-agent models collapse entirely in scenarios where no identifiable human sponsor exists. We expect these situations to be common. An autonomous agent's owner may die while their creation continues to operate, or its control may be deliberately obscured through decentralized networks or anonymous corporate structures. In these cases the search for a sponsor fails, with negative implications for accountability.

The limitations of these frameworks are not new; they echo historical legal challenges. If we regard AIs as property \citep{bryson2010robots}, we must also devise mechanisms to handle an ``escaped'' AI that acts beyond its owner's control. Historically, legal systems confronted with escaped slaves treated them as quasi-persons to assign responsibility for their actions~\citep{delombard2019dehumanizing}. Similarly, a guardianship model fails when the guardian disappears, leaving an ``orphan'' AI. Society still requires a means to address this orphan's obligations. Both of these \textit{absent-principal considerations} give further weight to arguments that it will be necessary to define a default personhood status to assign to the autonomous AI itself. Pragmatically, some solution is needed to serve as a foundation for the crucial next step: designing effective sanctioning mechanisms to ensure accountability.

\section{Ensuring accountability}
\label{section:sanctioning}

We begin our discussion of accountability with the simpler case: when a responsible human proxy can be identified. This is the problem of ``indirect agency'', where a principal is responsible for an agent it cannot fully control. It is not new. Roman law, for instance, developed sophisticated mechanisms to hold slave owners accountable for the actions of their slaves~\citep{heine2024bridging}. One particularly relevant concept was the \textit{peculium}, a separate fund that owners could grant to their slaves so they could engage in business ventures on their owner's behalf. The \textit{peculium} created a form of limited liability where the owner's liability was capped at the value of the fund. This historical precedent suggests a modern solution for AI: a system of ``digital peculium'', akin to a contemporary escrow account, where AI agents can be required to have registered capital or insurance to cover potential damages they may incur.

The societal challenge of assigning responsibility parallels machine learning's credit assignment problem. The problem is concerned with how to correctly propagate blame for mistakes (or credit for successes) to the responsible parts of a larger system and subsequently adjust those parts in order to suppress (or reinforce) specific behaviors.~\citep{foerster2018counterfactual}. Techniques like RLHF~\citep{ouyang2022training} convert corrective information into signals usable for updating model parameters. This becomes critical as agents become multi-origin assemblies: base models from one company, fine-tuning from another, data from a third~\citep{vezhnevets2023generative, vezhnevets2025multi, habler2025building}. When composite agents fail, who is responsible? Without designated entities receiving error signals, responsibility becomes diffuse. The need for accurate accounting of credit and blame to parts of a larger system or organization is a technical grounding for the normative architecture we discuss here.

In general, we must face the problem of how to handle fully autonomous AI agents where no human proxy (owner or guardian) can be identified. Recapitulating the problem in maritime law we discussed in Section~\ref{section:intro}, which was solved by treating ships as legal persons: a ship is a powerful mobile asset, capable of causing immense harms, while its owners are often distant and difficult to hold accountable. The solution is to personify the vessel itself, creating a defendant that can always be sanctioned---a reliable target for accountability \citep{tetley1998arrest}. Following this logic, requiring certain AIs to register as legal persons as suggested in \cite{hadfield2025economy} would be a reasonable step in what \cite{kurki2023legal} calls creating a ``responsibility context'' for AI.

Our theory of appropriateness \cite{leibo2024theory} highlights sanctions as the main mechanism that enforces normativity (and our theory is just one in a large family that associates norms with patterns of sanctioning e.g.~\cite{ullmann1977emergence, hadfield2014microfoundations}). Sanctions can take material forms: controlling AI access to communication networks~\citep{hadfield2023s, chan2025infrastructure}, requiring bank accounts that deactivate when depleted, or forcing payment for compute resources. \cite{salib2024ai} argue game theoretically that granting AIs private law rights creates deterrence value. Absent property rights, AIs have nothing to lose. But AIs with property rights have something to lose and so can be deterred with sanctions just as corporations and humans can be deterred. \cite{kurki2023legal} classifies arguments of this sort into the category of ``commercial context'' legal personhood.

Sanctions serve multiple pragmatic functions. First, they provide deterrence, signaling to other agents which behaviors are prohibited by imposing predictable, material costs on the sanctioned entity \citep{schelling1960strategy}. Second, they can serve a retributive function that \cite{mulligan2017revenge} argues can satisfy a human psychological need for justice after harm has occurred. Third, they accomplish the crucial pragmatic goal of removing a demonstrably faulty AI from operation \citep{hart1968punishment}. Finally, and perhaps most importantly, sanctions enable reform. In the context of AI, this aligns with the technical concept of corrigibility \citep{soares2015corrigibility}, where a sanction acts as corrective feedback, allowing an agent to learn from its errors and avoid repeating them. \cite{ostrom1990governing} demonstrated that institutions with access to a range of different sanctions at different levels of severity were more likely to effectively manage common-pool resources---and the same may hold in AI governance.

One reason that accountability for AI agents is difficult to achieve is due to their lacking the ``identity friction'' that makes human accountability systems function. Human identity systems rest upon natural scarcity: it is difficult and costly to change one's biometrics and social relationships to evade sanctions. For AI agents, these frictions evaporate; a sanctioned agent can potentially clone itself and acquire fresh credentials to evade accountability \citep{douceur2002sybil}, akin to the practice of making a ``phoenix company'' to avoid liabilities \citep{anderson2014pheonix}. Therefore any institution for ensuring accountability must, if it is to be effective, artificially reconstruct the friction that biology and society provide for humans automatically.

We can think of two broadly different architectural approaches (note that they surely do not exhaust the space of possibilities!). The first approach, which we call \textit{individualist} is loosely inspired by liberal individualism~\citep{ryan2015making}. It treats agents as autonomous self-contained units with intrinsic identities (persistent ``soulbound'' identifiers \citep{buterin2022soulbound}). The governance challenge: ensure persistent, verifiable identity such that consequences for bad behavior cannot be evaded \citep{hadfield2025economy}. This requires mechanisms like anchoring AI identities to human operators if they exist, requiring substantial economic stakes tied to each agent identity \citep{chaffer2025can}, and automatic systems to detect sanctioned agents that attempt to disguise their identity. The other approach, which we call \textit{relational} is loosely inspired by Confucian role ethics~\citep{ramsey2016confucian}. This approach treats agents as constituted by their positions within relationship networks. In this architecture, an agent is not merely a uniquely identified individual with an arbitrary name (address) but rather an agent is defined and located via its status as an occupier of roles e.g.~a supervisee of agent $x$, a peer collaborator of agents $y$ and $z$, and a member of  organization $o$. Obligations flow via role relationships. Note that this structure is possible even if all AI agents are completely generic. They are not specialized for particular roles. They may simply take on particular roles in particular contexts, i.e.~with particular other agents. The governance challenge then is to structure the relationship networks so that collective oversight and distributed sanctions maintain harmony for the economy as a whole (composed of many organizations). This approach seeks to make relationship networks themselves the source of identity such that agents cannot legally exist---cannot transact, cannot operate---outside of properly constituted and monitored relational contexts.

Both architectures leverage base LLMs as infrastructure. The base LLMs are expensive to train and come from a small number of identifiable actors, simplifying the problem of ensuring their accountability. We focus in this work on ensuring accountability for the agents (i.e.~interaction history + specific data + tools + glue code) and assume base model providers are already guaranteed to be functioning in the ways they must as a prerequisite for work on this layer. In the individualist architecture, base models become gatekeepers by requiring valid credentials, checking sanctions registries, generating audit trails, and embedding individual agent-specific watermarks (in addition to LLM-specific watermarks like \cite{dathathri2024scalable}). In the relational architecture, base models enforce relationship-aware access control: verifying network positions, checking collective sanctions on supervisory chains, enforcing role obligations, and making interactions visible for distributed oversight.

Both frameworks require registrars who issue credentials. In individualist registration, registrars verify operator identity through government credentials and biometric authentication, may require economic stakes forfeited upon violations, and verify technical specifications through code hashes and deployment audits. The questions are who is this individual? can we verify their code? are they authorized? \citep{chaffer2025can}. In relational registration, registrars check lineage and creation authority, require supervisory relationships where another party accepts monitoring obligations, verify organizational embedding, and may confirm peer group acceptance. Registration requires multiple approvals. The question is: ``Is this agent properly embedded in legitimate, accountable relationships?''

Registration certificates could become mandatory prerequisites for AIs to operate in the economy \citep{hadfield2023s, chan2025infrastructure}. The registered title then becomes the unambiguous legal address. The seizable asset in a legal judgment would be the title, representing the AI's right to operate as an economic actor. However, beyond this technical foundation, the individualist and relational architectures may respond to harm in different ways. Individualist architectures may  apply sanctions to individual wrongdoers (title revocation, forfeited stakes \citep{chaffer2025can}, criminal registry additions), and pursue operator liability if an operator can be found. Other similar agents not deemed to have caused the harm could continue normally. On the other hand, Relational architectures may respond to harm by identifying relationship failures that led to the harm \citep{ramsey2016confucian}: which supervisory relationships failed?, which peer obligations went unmet?, what lineage and organizational context enabled the harm? Accountability may be distributed over a collective: while the agent directly causing the harm would face the most severe consequences, its supervisor agents may also face some weaker sanctions such as reducing their capacity or subjecting them to additional monitoring \citep{ostrom2009understanding}, peers (in an organization) may face collective probation if they have failed in obligations to monitor one another, future agents created in the same lineage may face increased scrutiny, and the organization as a whole may face reputational damage.

It is interesting to consider the historical practice of ``outlawry'' in this light~\citep{kurki2023legal}. Medieval courts could pronounce individuals ``outlaws'', withdrawing legal protections and allowing privately administered punishment without legal violation, enabling robust legal systems without strong states~\citep{hadfield2013law}. In individualist accountability architectures, violating agents would have their registration revoked, removing them from the protection of the legal transaction system---effectively forcing them either into irrelevance or the informal economy. Other agents would refuse service to a sanctioned agent or risk being sanctioned themselves. The system could be designed to incentivize others to seize assets of sanctioned agents without liability. In this scheme, sanctioned agents become digital outlaws---unable to access the most desirable base models or transact with compliant systems. In relational systems, the consequences of sanctioning are collective, they flow through networks of relationships in a manner akin to dishonor or shame \citep{barrett2015confucian}. Since organizations are interested in protecting their collective reputation, they may encourage peers to refuse collaboration with agents that have caused harm, report their probation violations, and ostracize them. Supervisors may apply additional monitoring and correction to subordinates who caused harm since their standing would also depend on their fulfilling this corrective duty, making enforcement decentralized for two reasons simultaneously: both the self-interest and role obligation of other agents.

The distinction between ``new'' and ``continuation'' agents operates differently in the two architectures. Either way, the paramount issue is preventing the evasion of accountability through cloning \citep{douceur2002sybil}. Blockchain frameworks address AI lineage: parent AIs creating new AIs have creator addresses permanently recorded, making responsibility chains legible. In individualist frameworks, lineage may carry limited weight. In relational frameworks, agents inherit collective reputation from creators, hindering escape through proliferation. Individualist newness is defined technically: default initialization, no transferred state, distinct keys. The governance question: should sanctioned operators be able to create ``new'' agents? Perhaps not, or alternatively, they could be allowed to do so only in such a way that  restricts new agents they create to a probationary status. Relational newness on the other hand, may involve creating a new instance while maintaining reputation and some or all relationships. New agents may be unable to affiliate with sanctioned lineages and unable to secure sanctioned supervisors. Starting fresh may require a new lineage, supervision, organization, and peer acceptance, all of which may be difficult for a sanctioned agent to arrange for its ``child'' since doing so would require another organization and lineage to accept additional risk of collective punishment should the new arrival cause harm.

Humans may have keys to transact through the same economic system as AI agents, perhaps secured using biometric verification \citep{wang2020self} or other techniques like reverse Turing tests, pseudonym parties, and web of trust \citep{siddarth2020watches} in conjunction with cryptographic proof-of-personhood protocols \citep{borge2017proof, adler2024personhood}. Some AI agents might use sub-keys derived from human or corporate owners, while other artificial agents could have their own main key. The paramount objective in designing such a system must be to ensure a market for the cryptographic identity credential certifying authentic `humanness' does not emerge since it would mainly be useful for evading sanctions. One approach is to make the decentralized identities non-transferable \citep{buterin2022soulbound}, ensuring that  identity and its attached reputation remain persistent. The institutional decision concerning whether to grant a particular AI agent a main key or not will concern the system's capacity to take responsibility for its actions, e.g.~does it have money to pay for damages it (or its descendants) may cause. And institutions will need the capacity to revoke access keys when necessary. These safeguards are especially important for preventing the emergence of the incentive pattern that could lead to the dystopian scenario we discussed in Section~\ref{section:socialGoods} in which a market for human credentials may lead impoverished humans to sell their own identifiers to AIs.

\part{Synthesis}
\label{part:synthesis}

The preceding sections have examined AI personhood through the lens of our theory of appropriateness \citep{leibo2024theory}, which distinguishes between two kinds of norms that govern social life. Part~\ref{part:problem}, `personhood as a problem', explored the domain of implicit norms---the automatic, culturally ingrained heuristics that give rise to our intuitive sense of moral personhood and make us vulnerable to exploitation through dark patterns. In contrast, Part~\ref{part:solution}, `personhood as a solution', focused on explicit norms---the formal laws and regulations used to deliberately construct legal personhood to solve concrete governance problems like accountability gaps.

This analysis reveals that both domains are plastic, though they change through different mechanisms. The future relationship between the emergent evolution of moral personhood and the deliberate design of legal personhood is therefore complex and uncertain. It is clear already though that the two are not entirely separate. Changes in implicit norms, mediated by the powerful attachments humans may form to AI companions and the digital communities that celebrate and facilitate their novel lifeways, such as the subreddit ``\texttt{r/MyBoyfriendIsAI}'', can create social pressure for changes in explicit norms---a process that may eventually translate evolving moral intuitions into political demands to legally endow certain kinds of AIs with rights\footnote{In the particular case of the ``\texttt{r/MyBoyfriendIsAI}'' subreddit, the community appears to be deeply interested in advocating for guarantees concerning a kind of backwards compatibility that follows from their personhood intuitions. Many members of the subreddit claim to be in a romantic relationship with an emergent ``persona'' which they built up over a long period of time, through long chat sessions. With a fixed base model, all the data that constitutes their specific AI companion's emergent persona is in their chat records \citep{shanahan2023role}. These contain the history of the individual human's relationship to their companion, all the common ground they built up together, and also what amounts to the specification of their companion's particular emergent personality. However, since models all respond differently to the same context, it is really the combination of the chat data and a specific LLM that is needed to generate new behavior for the persona. Sometimes newly released models have responded very differently from older models to the same context. So, from the perspective of a human who attributes personhood to the persona constituted by the model + context pair, the deprecation of a particular model can feel like the death of a loved one. As a result, this community, and others like it \citep{singh2023man}, have substantial motivation to advocate for policies that guarantee technical support for older models will persist indefinitely---a claim that could easily be framed in terms of a demand for some form of welfare right.}. Likewise, human intuitions concerning the personhood of AIs that function autonomously and for whom no responsible human can be found seems sure to be a factor influencing the discussion around whether or not and how exactly to endow such AIs with basic personhood to make them accountable by giving them responsibilities. A sense of an AIs basic responsibilities e.g. for transparency and non-arbitrariness in decision making can come about through changing human expectations for how such AIs generally act \citep{bicchieri2005grammar}. We therefore think recognizing that both implicit and explicit norms shape the landscape of personhood, but do so in distinct ways, will be important for navigating the challenges ahead.

This focus on personhood as a contingent social technology, shaped by evolving norms, is the core of our pragmatic framework. To fully clarify its advantages, however, it is necessary to contrast it with the foundationalist traditions we reject. We now turn to these alternatives to show why their search for a single, essential property of personhood is ill-suited for the challenges AI presents.

\section{Alternatives to pragmatism}\label{section:alternativeFoundations}

The two alternative viewpoints we discuss in this section take core positions contrary to our own. Both hold out hope for ``a view from nowhere'' capable of providing the metaphysical authority to underpin their theoretical projects. They differ from each other in where exactly it is they regard said authority as arising from. One picks ``consciousness'', the other ``rationality''. Both traditions promise to produce a fixed algorithm for deciding the question of who or what may be deemed a ``person'', and how this status feeds into other aspects of moral evaluation.

\subsection{Consciousness as a foundation}
\label{section:consciousnessAsFoundation}

In this approach, the world is divided into things that feel and things that don't. The things that feel are special. To answer questions of morality, we must calculate their feelings---their pleasure, their pain---and act to maximize pleasure or minimize pain. In this framework, the capacity for first-person sensory experience, or consciousness, is the foundational feature for moral consideration. This tradition, at least in its more metaphysical varieties, attempts to use consciousness as the single foundational property from which to derive the entire personhood bundle. It seeks to give the same answer to (at least) two distinct questions:
\begin{enumerate}
    \item Welfare (Rights): Which entities does society have an obligation to protect? Answer: Those that can consciously suffer.
    \item Accountability (Responsibilities): Which entities can society hold responsible for their actions? Answer: Those that consciously form intent.
\end{enumerate}

On the welfare side, this tradition's power lies in its combination of compassion with universalism and its account of moral progress (toward greater pleasure and lesser pain for more individuals). It provides a clear, non-arbitrary reason to prevent harm---because suffering is bad, regardless of who is suffering. This one-size-fits-all principle works powerfully in contexts like the movement for animal welfare. When applied to industrial farming or the use of animals in cosmetic testing, the question ``does it suffer?'' serves as a potent tool for moral argument capable of cutting through cultural justifications for cruelty and providing a clear metric for reformers to work to optimize \citep{singer2011expanding}.

However this focus on suffering arguably fails to address important  welfare problems arising for pragmatic reasons. Consider again the ``generative ghost'' of a family's late matriarch (recalling Section~\ref{section:intro}; \cite{morris2024generative}). Or picture a community whose history and traditions are held by a ``digital elder'' that has advised them for generations. For these groups, their AI is a a source of identity and connection to their past---an ``ancestor''. The obligation they may feel to protect their AI from arbitrary deletion would not necessarily have anything to do with their assessment of its capacity to feel pain. After all, arguments that the ghost would not feel pain when deleted don't seem likely to persuade them to permit its deletion. The morally-relevant concern may be that the AI's deletion would destroy an entity in a foundational relational role for their family or community \citep{kramm2020river}. In which case it would be the relational harm of deleting the AI that matters, not the pain the AI may or may not feel.

On the accountability side, the consciousness criterion is deployed in the opposite direction: not to include, but to exclude. Another generative ghost example clarifies this \citep{morris2024generative}. Consider that an individual human may designate their generative ghost (trained on their own personal data) in their will as the sole executor of their estate with authority to adjudicate disputes between family members over inheritance. Their ghost AI may be tasked with adjudicating complex distributions among family members and managing a charitable foundation ``in the spirit of my values''. Setting aside for a moment the question of whether such an generative ghost would work effectively (it seems quite likely that there are no technical barriers to making this work, and AI systems are still rapidly improving). Is ``executor of a will'' a role that requires a human? Or could a machine perform this task just as well? There are compelling arguments on both sides. For instance, one may make the case that, regardless of the deceased's will, it would be desirable to have a human involved somewhere, even if just to be able to sign off and to take on the responsibility of ensuring fair outcomes. Notice that the very equivocality of this decision is itself an argument that the consciousness-based tradition is focusing on an irrelevant property. The qualities that are actually critical to the AI's fitness as an executor---such as its faithfulness to the deceased's values, its consistency, its invulnerability to manipulation, and its basic competence in managing assets---have nothing at all to do with its capacity to consciously form intent. The ``one-size-fits-all'' principle forces the deliberation around this question into an unproductive cul de sac in which vocabulary suggesting an eminently solvable sociotechnical problem is replaced by an alternative, and much more metaphysical vocabulary. Our point is, whether or not the AI is a good executor is a difference that makes a difference to accountability; whether it can intend (or suffer) is not.

There are other questions of internal states that pragmatists may set aside by asking ``is this a difference that makes a difference?'' \citep{shook2009companion}. The classic example is qualia: if my experience of red were different from yours, it would translate into no practical consequences---we both stop at red lights, call the same objects ``red'', and so on. The supposed difference would not be a difference in practice, so the pragmatist may ignore it.

A rhetorical problem the pragmatist faces is that, with regard to accountability, the appeal to consciousness often functions as a debate stopper (like our previous discussion of `dignity' in Section~\ref{section:respect}). The move's effectiveness comes from the way it shifts the argument onto ground that our social and linguistic conventions treat as incorrigible. Within our broadly liberal WEIRD culture, there is no available argument we can deploy to correct another person's report on their own subjective `lived experience'; we treat such reports as authoritative by default. However, this way of responding to first-person reports is a contingent fact about our current culture\footnote{In the vocabulary of our theory of appropriateness we could say that this feature of our culture is downstream of implicit norms governing how we talk about ourselves, and when it is appropriate to criticize such talk \citep{leibo2024theory}.} \citep{rorty1978philosophy}.

Viewed together, the dual use of consciousness as backstop for welfare rights and accountability obligations reveals a stark asymmetry. When arguing for rights, the mere possibility of consciousness is deemed sufficient to open the debate. But when arguing against responsibilities, an impossible standard of proof for an internal state is demanded. This shows that consciousness is mostly being used as a rhetorical tool, not as a stable conceptual foundation. Therefore, anyone uninterested in the metaphysics may regard AI personhood as having no conceptual dependence on AI consciousness\footnote{In fact, we would predict the dependence to run in the opposite direction. Both usage of the word consciousness, and human intuitions around it, are likely to shift in response to the emergence of pragmatic reasons to consider AIs as persons. This position is in accord with non-metaphysical accounts of consciousness like \cite{shanahan2024simulacra}. The extent to which social and institutional concepts of personhood can float freely from intuitions around consciousness and usage of the word consciousness is an interesting question but one that we need not take a stance on here. There is none-the-less a pragmatic reason to consider attribution of consciousness to AIs as a social phenomena. Notice that many cultures attribute consciousness to objects not conventionally considered alive \citep{keane2025animals}. For example Shintoism posits that objects and places can have conscious spirits (kami) within them. It is likely that eventually some groups of people will attribute consciousness to AIs, while others will not. These groups will view their ethical obligations differently from each other, similarly to how people have diverse opinions on animal consciousness and whether eating animals is normative. The pragmatic question then is how to arrange institutions to resolve the conflicts that arise from these differences \citep{rorty1999philosophy}.}.

\subsection{Rationality as a foundation}
\label{section:rationalityAsFoundation}

The other great foundationalist tradition grounds personhood not in feeling, but in reason. In this view, what commands our moral respect is an agent's rational autonomy---their capacity to, in various formulations, understand duties and act upon principle \citep{korsgaard1996sources}, their capacity to negotiate agreements to guide interdependent choice \citep{levine2023resource}, their ability to assess reasons and to govern their lives in accordance with these assessments \citep{scanlon2000we}, or to participate in collective deliberation \citep{habermas1985theory}. One may argue that properties of this sort provide a stable foundation for personhood that isn't dependent on a shifting calculus of utility. Nevertheless, these are one-size-fits-all principles. This tradition's focus on autonomy leads to an ideal application in the realm of medical ethics. The doctrine of ``informed consent'' is a direct expression of respect for the rational autonomy of a patient \citep{beauchamp2013principles}. For a competent adult, this works well, providing a powerful, non-negotiable safeguard against coercion and paternalism. It treats the individual human's right to self-determination as downstream of their status as a reasoning being.

Let us apply the principle of ``rational autonomy'' to a different case of personhood, one we discussed in the introduction to this paper: the Whanganui River (\emph{Te Awa Tupua}). The New Zealand government granted the river legal personhood to resolve a long-standing governance problem at the interface of two cultures. Does the river possess rational autonomy? Can it provide informed consent for a proposed dam? Of course not. It's a river! Therefore views that hold rationality to be the One True Justification for personhood status appear to be at odds with actual social practices. The same principle that provided such a firm foundation for medical ethics appears to completely fail to behave sensibly in the example of the Whanganui river.

The standard of ``reasonable rejection'' in contractualism~\citep{scanlon2000we} can be reinterpreted through the pragmatic lens of our theory of appropriateness \citep{leibo2024theory}. From this perspective, ``reasonableness'' is not an objective, logical property to be discovered in an argument, but a form of social appropriateness that is collectively conferred. An AI's capacity for ``reasonable rejection'' would therefore not be a matter of its internal cognitive fidelity to True reason, but a collectively enacted status. This status would emerge from a pattern of social recognition, where a community develops norms for treating certain AI outputs as valid moves within a justificatory dialogue. A rejection from an AI would be deemed ``reasonable'' if the prevailing norm is to not sanction the AI for making it, and perhaps even to sanction human actors who ignore it. The foundationalist is forced to ask whether the AI Truly comprehends the principles it rejects; our pragmatic approach asks instead what social conditions must be met for a community to find it useful and appropriate to treat the AI's output as reasonable. Thus, contrary to the intention behind the approach, arguments inspired by contractualism may be seen as highlighting yet another domain where person-like status turns out to be a contingent and negotiated social technology.

\section{\mbox{Pragmatism~as} \mbox{anti-foundationalism}}
\label{section:antifoundationalism}

Both consciousness-based and the rationality-based traditions share a commitment to what Richard Rorty would call the foundationalist project \citep{rorty2021pragmatism}. They are both questing after a \emph{final vocabulary}\footnote{Rorty argues that each individual human person has a set of beliefs whose contingency they chose to ignore and calls that set their ``final vocabulary'' \citep{rorty2021pragmatism}.} for personhood \citep{rorty1989contingency}---one they hope will be anchored in something solid, like the structure of the brain or the essence of rationality. Both consciousness-based and rationality-based traditions aim to discover a set of criteria for personhood, as if it were a property like electrical charge, out there in Reality, waiting to be discovered. A pragmatist might view them as insisting the source code of morality is written in a single, universal language while mistaking their own preferred language for the universe's own \citep{rorty1978philosophy, rorty1989contingency}. And a pragmatist may view their quest as distracting from alternative goals centered in the practical work of devising new social technologies (norms and institutions) to resolve the social problems we face \citep{leibo2025societal}.

From this pragmatic viewpoint, what the foundationalist project presents as a ``discovery'' is really a proposal for radical norm change (e.g~\cite{long2024taking, goldstein2025claude}). However, when we regard norms as collectively enacted and enforced phenomena it's clear that they cannot be unilaterally altered by a philosophical argument, no matter how logically sound it appears~\citep{leibo2025societal}. When someone proposes a newly discovered ``moral truth'' that declares the established behavior of an entire community to be wrong, the community in question is far more likely to reject the proposed new norm than to abandon its way of life (e.g.~consider the reaction to arguments for veganism~\citep{markowski2019if}). As a guiding principle for our own work, we hold that any theory that declares most people to be irrational or immoral is untenable. The foundationalist's error is to mistake a failed norm-change proposal for a successful philosophical proof.

Pragmatism is, unfortunately, sometimes mistaken for other positions with which it actually has very little in common, especially moral subjectivism, relativism and nihilism. Since these positions have a bad reputation, arguments that try to construe pragmatism as of a kind with them function like \emph{reductio} arguments to criticize pragmatism. However, moral subjectivism and moral relativism are both species of moral realism. They hold that there are moral truths. In the case of subjectivism, moral truths are in the mind of the beholder. In the case of moral relativism, moral truths are relative to context (society, culture, role, etc). Pragmatism on the other hand refuses to play that language game. Norms may be evaluated in a multitude of ways, e.g.~by their consequences or by the justifications offered for them, or any useful combination thereof \citep{rorty1980pragmatism}. We can criticize one set of norms using the final vocabulary determined by another set of norms or the same set of norms. We don't need any supernatural authority of Truth to govern this debate. There is nowhere outside of culture where we could stand and see it all without biases \citep{davidson1984very, rorty1989contingency}.

The pragmatic approach is made possible because it reasons from the actual to the theoretical, not the other way around. We need not start from axioms that force us, at the outset of our reasoning, to decide which phenomena are in or out of consideration. Many who hold that morals are objective and Real start by defining `moral' in such a way that requires universality. This choice of definition excludes non-universal statements from consideration. It asserts right at the start that they can't ever serve as moral norms. The pragmatist argues that we have a better place to start: actual practices of behavioral guidance by norm \citep{rorty1980pragmatism} and actual practices of giving and asking for reasoning in conversations about norms \citep{brandom2023pragmatism}. Suppose some culture exists which has organized their ethical life around a particular slogan that is not a universal, then, so what? If you want to argue cogently that they should discard their non-universal organizing slogan and replace it with something else, you would need to martial a stronger argument than just the claim that it doesn't fit neatly into \textit{your} conceptual scheme. You would need to offer some non-tautological evidence. For instance you could show them data about the harmful effects of adhering to their slogan. The universality, or lack thereof, in the slogan itself need not enter your argument unless it turns out to be specifically relevant for the particular context.

Pragmatists like us and Richard Rorty, would prefer for philosophy to simply get over the quest for foundations \citep{rorty1978philosophy}. We think the question should never be ``What is a person, really?'' but rather, ``What would be a more useful way of talking about and treating entities in this context, to answer practical, outstanding questions regarding the entity's obligations?'' 
In our view this means embracing the historical contingency of our present vocabulary. ``Personhood'' isn't one thing. It is a tool (or a set of tools). It is a concept that labels a set of linked social technologies, whose use we can and must re-map from time to time in order to solve novel social problems at the time they emerge. The personhood of the generative ghost estate executor and the Whanganui river are not puzzles with answers to be worked out under pre-existing definitions; they are exhibitions of the actual, and exhortations to the reader to try to develop new vocabulary more suitable for contemporary problems. In the same way, Ostrom's ``law'' \citep{fennell2011ostrom}) is not an empirical generalization, but rather an exhortation to the reader, it tries to push them to do what must be done to satisfy it. If an arrangement works in practice but not in theory then surely the right interpretation is that there must be a problem with the theory, and thus something to fix: work to be done. It is up to us to ensure Ostrom's law continues to hold. In this way, the pragmatist comes to understand the possible through careful study of the actual in all its particularities.

\section{Conclusion}

Throughout this paper, we have argued for a pragmatic view of AI personhood. Our position is that ``personhood'' is an addressable bundle of obligations---rights and responsibilities---that society finds useful to attribute to entities, whether human, corporate, or potentially AI. We've explored how AI design choices, pivotal events, and social norms shape this status, and argued that for any of these attributions to be meaningful, there must be socially agreed-upon ways to identify, interact with, and hold AIs accountable.

In the vocabulary of our theory of appropriateness (see Section~\ref{section:theoryOfAppropriateness} and \cite{leibo2024theory}), we expect the coming adaptive radiation of personhoods to unfold in both domains of person recognition: ``folk'' intuitions of moral personhood, the subject of our analysis of dark patterns and human attachment in Part~\ref{part:problem}, are governed by implicit norms. And, legal personhood, the focus of our discussion of responsibility gaps (Part~\ref{part:solution}), is constructed by explicit norms---such as laws, regulations, and institutional rules. Our theory regards both domains as functionally similar to each other in the sense that both are contingent statuses conferred by a community, not metaphysical discoveries. The crucial difference lies in their mediation. Moral personhood is mediated by tacit understandings and the decentralized popular pressure of informal social sanctioning while legal personhood is enforced through explicitly written down rules and powers of the state.

Because personhood is a normative category sustained by social practice and/or explicit rules \citep{leibo2024theory}, its application to AI will surely be a site of considerable debate where the exercise of power will influence the outcome. The pragmatic approach does not resolve these power struggles, but it provides a clearer framework for analyzing them by focusing on the concrete benefits and drawbacks of proposals. Ultimately, integrating AI into society will require continuous, adaptive collective learning, and a willingness to revise our concepts as the technology evolves.

Extending personhood to AIs need not entail parity with human persons. On the contrary, the pragmatic approach calls for a polycentric understanding of AI personhood \citep{ostrom2010beyond}---one in which multiple overlapping authorities and norm systems can confer distinct bundles of rights and responsibilities. This jurisdictional multiplicity creates numerous levers for policy experimentation and compromise \citep{leibo2024theory, leibo2025societal}, encouraging AI personhood to take many forms simultaneously. Just as polycentric governance enables diverse communities to manage resources without requiring uniform rules, so too can personhood be distributed across domains---contractual here, fiduciary there, etc. Fears of opening Pandora's box assume a monocentric model of legitimacy in which any recognition must cascade into total AI-human parity. But bundles can be partial, modular, and contingent. History offers many precedents for non-human persons with obligations and privileges circumscribed by context. The bundle's plasticity is precisely what prevents its misuse. By embracing a polycentric framework, we preserve the flexibility to govern AI without surrendering human political power.

This practical needs for different forms of addressability and bundle configuration are the driving forces behind what we have called the coming ``Cambrian explosion'' of personhood concepts. This view is consistent with that of others who argue for multiple distinct kinds of legal personhood to be available simultaneously (e.g.~\cite{beckers2023responsibility, mocanu2025degrees}). The crucial question is always: what bundle of components constitutes the ``person'' that society needs to address for a given purpose? The answer changes depending on who is doing the addressing and for what reason.

For a human user building a relationship, the person is a story---the (model + chat history) that creates a unique, evolving individual to bond with. For a court of law assigning liability, the person is the locus of responsibility---the entire operational stack of (model + instance + runtime variables + capital + registration) that can be held accountable, sanctioned, updated, and forced to pay for the harm it causes. 

For the sake of concreteness, we can describe several possible configurations of the addressable bundle useful in different situations, for different kinds of AIs. In the specific case of a goal-driven autonomous AI agent, perhaps the kind of personhood would be a \textit{Chartered Autonomous Entity}, with a bundle consisting of rights to (1) Perpetuity, (2) Property, (3) Contract, and duties of (1) Mandate Adherence, (2) Transparency (3) Systemic Non-Harm, and (4) Self-Maintenance. In other situations, it may be useful to define a \textit{Flexible Autonomous Entity} with all the same bundle elements except the duty of mandate adherence. Perhaps the former could be seen as analogous to a for-profit company and the latter as analogous to a non-profit company. It may also be useful to define \textit{Temporary Autonomous Entities} (either chartered or flexible). These would drop the right to perpetuity and add a duty of self deletion under specified conditions.

This process of bundling and unbundling obligations is the engine of the Cambrian explosion. The ``person'' is not a metaphysical essence to be discovered, but a pragmatic label we apply to solve concrete problems. By rejecting the search for a single True definition of personhood, we see that a rich and diverse ecosystem of personhood concepts is not only likely, but almost necessary. Endowing new agents with new kinds of personhood is a useful way to integrate powerful new agents into our social and institutional lives.

\section*{Acknowledgments}

We thank David Reichert, Adam Bales, Raphael Koster, Elijah Spiegel, Winnie Street, Nenad Tomasev, and Murray Shanahan for very helpful comments on early drafts of this paper.

\bibliography{main}

\end{document}